\documentclass[10pt,a4paper,fontset=none]{ctexart}

\usepackage[a4paper,left=19mm,right=19mm,top=18mm,bottom=19mm,headheight=14pt]{geometry}
\usepackage{fontspec}
\usepackage{xeCJK}
\usepackage{amsmath}
\usepackage{unicode-math}
\usepackage{booktabs,longtable,tabularx,array}
\usepackage{enumitem}
\usepackage{xcolor}
\usepackage{tikz}
\usetikzlibrary{arrows.meta,positioning,fit,calc}
\usepackage{caption}
\usepackage{titlesec}
\usepackage{fancyhdr}
\usepackage{microtype}
\usepackage{hyperref}
\usepackage{bookmark}
\renewcommand{\abstractname}{Abstract}
\renewcommand{\contentsname}{Contents}
\renewcommand{\figurename}{Figure}
\renewcommand{\tablename}{Table}
\renewcommand{\refname}{References}
\renewcommand{\appendixname}{Appendix}

\definecolor{DeepBlue}{HTML}{17365D}
\definecolor{Accent}{HTML}{2459A6}
\definecolor{SoftBlue}{HTML}{EAF1FB}
\definecolor{SoftGray}{HTML}{F4F6F8}
\definecolor{LineGray}{HTML}{D7DFEA}
\hypersetup{
  colorlinks=true,
  linkcolor=DeepBlue,
  citecolor=Accent,
  urlcolor=Accent,
  bookmarksnumbered=true,
pdftitle={The Evolution of Mixture-of-Experts Architectures in Large Language Models: Routing, Topology, Load Balancing, and Expert Parallelism},
  pdfauthor={Jiguo Li},
  pdfsubject={Large Language Model Mixture-of-Experts Architecture Survey},
  pdfkeywords={MoE, Routing, Load Balancing, Expert Parallelism, ScMoE}
}

\setlist[itemize]{leftmargin=1.8em,itemsep=1pt,topsep=2pt,parsep=0pt}
\setlist[enumerate]{leftmargin=2em,itemsep=1pt,topsep=2pt,parsep=0pt}

\titleformat{\section}{\Large\bfseries\color{DeepBlue}}{\thesection}{0.7em}{}
\titleformat{\subsection}{\large\bfseries\color{DeepBlue}}{\thesubsection}{0.65em}{}
\titleformat{\subsubsection}{\normalsize\bfseries\color{DeepBlue}}{\thesubsubsection}{0.55em}{}
\titlespacing*{\section}{0pt}{14pt}{5pt}
\titlespacing*{\subsection}{0pt}{10pt}{3pt}
\titlespacing*{\subsubsection}{0pt}{7pt}{2pt}

\newcolumntype{Y}{>{\raggedright\arraybackslash}X}
\newcolumntype{C}[1]{>{\centering\arraybackslash}p{#1}}
\newcommand{\term}[1]{\textbf{#1}}

\newcommand{\reported}{\textbf{ Paper report: }}
\newcommand{\inference}{\textbf{ This article's judgment: }}
\newcommand{\noteBox}[1]{%
  \begin{center}
  \fcolorbox{Accent}{SoftBlue}{\parbox{0.93\linewidth}{#1}}
  \end{center}}
\setCJKsansfont[AutoFakeBold=2]{gbsn00lp.ttf}
\setCJKmonofont{gbsn00lp.ttf}
\makeatletter
\let\BilingualOriginalAddContentsLine\addcontentsline
\def\BilingualTocExtension{entoc}
\renewcommand{\addcontentsline}[3]{%
  \def\BilingualRequestedExtension{#1}%
  \def\BilingualStandardToc{toc}%
  \ifx\BilingualRequestedExtension\BilingualStandardToc
    \BilingualOriginalAddContentsLine{\BilingualTocExtension}{#2}{#3}%
  \else
    \BilingualOriginalAddContentsLine{#1}{#2}{#3}%
  \fi
}
\renewcommand{\tableofcontents}{%
  \section*{\contentsname
    \@mkboth{\MakeUppercase\contentsname}{\MakeUppercase\contentsname}}%
  \@starttoc{\BilingualTocExtension}%
}
\newcommand{\BilingualUseEnglishToc}{\gdef\BilingualTocExtension{entoc}}
\newcommand{\BilingualUseChineseToc}{\gdef\BilingualTocExtension{zhtoc}}
\def\BilingualAnchorPrefix{en}
\let\BilingualOriginalAppendix\appendix
\renewcommand{\appendix}{%
  \BilingualOriginalAppendix
  \renewcommand*{\theHsection}{\BilingualAnchorPrefix.appendix.\Alph{section}}%
}
\makeatother

\begin{document}
\renewcommand{\abstractname}{Abstract}
\renewcommand{\contentsname}{Contents}
\renewcommand{\figurename}{Figure}
\renewcommand{\tablename}{Table}
\renewcommand{\refname}{References}
\renewcommand{\appendixname}{Appendix}
\BilingualUseEnglishToc
\renewcommand*{\theHsection}{en.\arabic{section}}
\renewcommand*{\theHsubsection}{en.\arabic{section}.\arabic{subsection}}
\renewcommand*{\theHsubsubsection}{en.\arabic{section}.\arabic{subsection}.\arabic{subsubsection}}
\renewcommand*{\theHequation}{en.\arabic{equation}}
\renewcommand*{\theHfigure}{en.\arabic{figure}}
\renewcommand*{\theHtable}{en.\arabic{table}}
\gdef\BilingualAnchorPrefix{en}
\markboth{}{}
\pdfbookmark[0]{English Version}{bilingual.english}
\begin{center}
  {\zihao{1}\bfseries\color{DeepBlue}
The Evolution of Mixture-of-Experts Architectures in Large Language Models}\par
  \vspace{4pt}
{\zihao{3}\color{DeepBlue} Routing, Topology, Load Balancing, and Expert Parallelism}\par
  \vspace{10pt}
{\normalsize\textbf{Jiguo Li\footnote{This report was completed with the assistance of Codex.}}}\par
  \vspace{2pt}
  {\small\href{mailto:jiguolee@gmail.com}{jiguolee@gmail.com}}\par
  \vspace{6pt}
{\small August 2026}\par
\end{center}
\vspace{5pt}
\hrule

\begin{abstract}
Mixture-of-Experts models increase parameter capacity while keeping the computation activated by each token bounded, but their architectural evolution cannot be explained by a chronological list of model releases alone. This technical survey synthesizes primary papers, official technical reports, and prior surveys to organize modern Mixture-of-Experts systems along five coupled dimensions: expert granularity, expert topology, routing freedom, the scope of load balancing, and execution structure. We describe eight architectural milestones as a dependency graph with six mainline developments and two orthogonal branches, rather than as eight successive generations. We then analyze individual systems through four control planes: Expert Topology, Routing, Balance, and Expert Parallelism. These planes specify which experts exist, which experts process each token, how aggregate load is controlled, and how selected computation is mapped onto physical devices. The framework connects algorithmic choices such as Top-k routing, shared experts, fine-grained experts, and dynamic expert composition with systems concerns including token dispatch, device placement, all-to-all communication, and communication-computation overlap. We conclude with equal-budget pretraining experiments, quality and systems metrics, and open research questions. The main trend is a shift from merely activating more sparse parameters toward decoupling semantic routing, computational budgets, and physical execution.
\end{abstract}


\noteBox{\textbf{Key point:} The eight milestones describe historical bottleneck migration, whereas the four control planes provide a structural view of an individual MoE system. They are complementary views, not two competing stage taxonomies.}

\clearpage
\begingroup
\small
\setlength{\parskip}{0pt}
\linespread{1.02}\selectfont
\setcounter{tocdepth}{2}
\tableofcontents
\endgroup
\clearpage

\section{Problem Definition and Analysis Scope}

Classic MoE research asks how a gating network allocates examples to local experts. Sparse MoE for large language models adds a stricter requirement: total parameter capacity should grow without a proportional increase in the parameters and computation activated by each token. Existing surveys provide complementary perspectives: Cai et al. organize algorithms, systems, and applications into a full-stack taxonomy\cite{en:cai2024survey}; Liu et al. analyze model, system, and hardware optimizations for inference\cite{en:liu2024inference}; and Zhu et al. place MoE within a broader landscape of sparse attention, state-space models, and hybrid architectures\cite{en:zhu2025speed}. Rather than repeating a model-by-model chronology, this report asks a structural question: \textbf{which bottleneck does each architectural change remove, and where does the bottleneck move next?}

The scope of the discussion is limited to sparse MoE in decoder-only LLM pre-training, focusing on covering FFN MoE. Expert Parallel (EP), All-to-All, token capacity and expert placement, which will adversely affect structure selection, are also discussed. Multimodal MoE, MoE-LoRA, external model ensemble and pure post-training expert fusion are outside the scope of this article. Model capability numbers are only used to illustrate architecture scalability and cannot be used as causal comparisons across papers: training data, number of tokens, optimizers, context length, post-training and evaluation pollution control are often not consistent.

\section{Unified formalization: What does MoE simultaneously optimize?}

\subsection{Token-choice MoE}

Given the hidden representation $x_t\in\mathbf{R}^{d}$ of the $t$th token, Router calculates expert affinity:
\begin{equation}
  s_{i,t}=\phi(x_t,e_i),\qquad
  \mathcal{K}_t=\operatorname{TopK}_{i\in\{1,\ldots,N\}}(s_{i,t}),
  \label{en:eq:router}
\end{equation}
Among them, $N$ is the number of routed experts, and $e_i$ is the expert routing embedding or Router weight. The output is
\begin{equation}
  y_t=x_t+\sum_{i\in\mathcal{K}_t}g_{i,t}E_i(x_t),
  \qquad
  g_{i,t}=\frac{\exp(s_{i,t})}{\sum_{j\in\mathcal{K}_t}\exp(s_{j,t})}.
  \label{en:eq:moe}
\end{equation}
If there are always-on shared experts, add $\sum_j E^{\mathrm{shared}}_j(x_t)$ to the formula ~\eqref{en:eq:moe}. From a formula perspective, MoE is just a sparse function combination; from a system perspective, $\mathcal{K}_t$ determines which devices the token spans, how many tokens each card gets, and the shape of each local GEMM.

\subsection{Three goals of capacity, quality and system efficiency}

MoE is not a single-objective optimization. Roughly speaking, the model hopes to maximize the total capacity $P_{\mathrm{total}}$ while controlling the single token activation amount $P_{\mathrm{active}}$:
\begin{equation}
  P_{\mathrm{active}} \approx P_{\mathrm{dense}}
  +\frac{k}{N}P_{\mathrm{routed}}+P_{\mathrm{shared}},
  \label{en:eq:active}
\end{equation}
However, the formula ~\eqref{en:eq:active} is only an approximation of the parameter caliber: attention, embedding, shared experts, different expert widths and frame statistics methods will all cause deviations. On the other hand, let the number of tokens received by expert $i$ in a statistical window be $n_i$, then the load variation coefficient is
\begin{equation}
  \operatorname{CV}_{\mathrm{expert}}
  =\frac{\sqrt{\frac{1}{N}\sum_i(n_i-\bar n)^2}}{\bar n}.
  \label{en:eq:cv}
\end{equation}
Low CV benefits Expert Parallel throughput but does not necessarily indicate useful knowledge specialization. The architectural tension is that \term{semantic specialization may be uneven, while physical execution must avoid severe stragglers}.

\begin{table}[htbp]
\centering
\caption{Six questions that determine MoE architectural differences}
\label{en:tab:sixquestions}
\small
\begin{tabularx}{\textwidth}{p{0.25\textwidth}Y}
\toprule
Design Issues & Typical Choices \\
\midrule
Who chooses whom & token-choice, expert-choice, balanced assignment, soft merging \\
How many experts are activated & fixed Top-1/2/8, threshold, adaptive-$k$, zero-compute slot \\
Expert size & coarse homogeneous, fine-grained, tiny, heterogeneous size \\
Is there a public path & pure routed, shared expert, dense residual, cross-layer shortcut \\
In which range to balance & sequence, micro-batch, global batch, device, node, communication \\
Who handles runtime dynamics & auxiliary loss, router bias, capacity/drop, placement, structural guarantees \\
\bottomrule
\end{tabularx}
\end{table}

\subsection{Four control planes: object, decision, control and execution}

The design questions in Table~\ref{en:tab:sixquestions} operate at different levels. To separate historical architectural evolution from the internal operation of one MoE system, we organize the system cross-section by \term{state variables, decision granularity, and update timescale} into four control planes. Topology is a slowly changing model structure; Routing makes discrete token-level decisions; Balance aggregates statistics and applies feedback over groups of tokens; and Expert Parallelism (EP) realizes logical decisions as communication and kernels on physical devices. Chapter 3 follows bottleneck migration over time, whereas Chapters 4--7 analyze these four control planes. \textbf{The control planes are not four additional stages of evolution.}

Formally, the four-level relationship can be written as
\begin{align}
  \mathcal{E} &= T(\theta_{\mathrm{topo}}),
  &&\text{Topology: construct the expert set, groups, and sharing relations}; \notag\\
  \mathcal{K}_t &= R(x_t,\mathcal{E};\theta_{\mathrm{route}},b),
  &&\text{Routing: select an expert subset for token }t; \notag\\
  n_i &= \sum_t \mathbf{1}[i\in\mathcal{K}_t],\quad
  (b,\alpha,c)\leftarrow C(\{n_i\},\pi),
  &&\text{Balance: update bias, loss, or capacity from aggregate load}; \notag\\
  y_t &= \operatorname{EP}\!\left(x_t,\mathcal{K}_t,\pi,\sigma\right),
  &&\text{EP: execute dispatch and combine under placement }\pi\text{ and schedule }\sigma.
  \label{en:eq:fourplanes}
\end{align}
Among them, $\mathcal{E}$ is the logical expert set, $\mathcal{K}_t$ is the routing result of the token, $n_i$ is the load of expert $i$ in the statistics window, $b$, $\alpha$, and $c$ represent Router bias, auxiliary loss intensity and capacity respectively. Control volume, $\pi$ is the mapping from expert to device, and $\sigma$ is communication and computing scheduling. Formula \eqref{en:eq:fourplanes} explains: The four layers are not independent modules, but a closed loop.

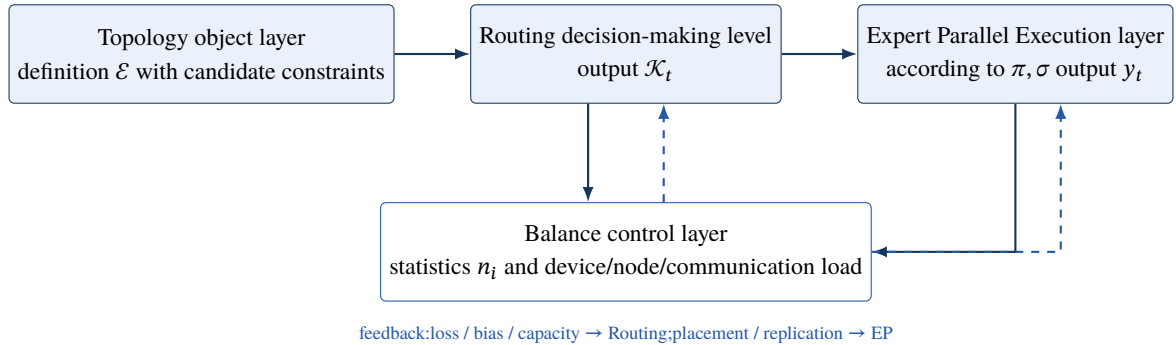
\begin{figure}[htbp]
\centering
\begin{tikzpicture}[
  node distance=10mm and 10mm,
  plane/.style={draw=DeepBlue,rounded corners=2pt,fill=SoftBlue,minimum width=38mm,minimum height=13mm,align=center,font=\small},
  control/.style={draw=Accent,rounded corners=2pt,fill=white,minimum width=58mm,minimum height=13mm,align=center,font=\small},
  arr/.style={-{Latex[length=2mm]},thick,draw=DeepBlue},
  fb/.style={-{Latex[length=2mm]},thick,dashed,draw=Accent}
]
\node[plane] (topo) {Topology object layer\\definition $\mathcal{E}$ with candidate constraints};
\node[plane,right=of topo] (route) {Routing decision-making level\\output $\mathcal{K}_t$};
\node[plane,right=of route] (ep) {Expert Parallel Execution layer\\according to $\pi,\sigma$ output $y_t$};
\node[control,below=13mm of route] (bal) {Balance control layer\\statistics $n_i$ and device/node/communication load};
\draw[arr] (topo)--(route);
\draw[arr] (route)--(ep);
\draw[arr] ([xshift=-5mm]route.south)--([xshift=-5mm]bal.north);
\draw[fb] ([xshift=5mm]bal.north)--([xshift=5mm]route.south);
\draw[arr] (ep.south) |- (bal.east);
\draw[fb] (bal.east) -| ([xshift=6mm]ep.south);
\node[font=\scriptsize\color{Accent},below=2mm of bal] {feedback:loss / bias / capacity $\rightarrow$ Routing;placement / replication $\rightarrow$ EP};
\end{tikzpicture}
\caption{The closed-loop relationship between Topology, Routing, Balance and Expert Parallel. The first three solid lines form the forward execution chain, and the dashed lines represent the feedback of statistics and system costs on routing and placement.}
\label{en:fig:fourplanes}
\end{figure}

\begin{table}[htbp]
\centering
\caption{Criteria for the division of the four control planes}
\label{en:tab:fourplanes}
\small
\begin{tabularx}{\textwidth}{>{\raggedright\arraybackslash}p{0.17\textwidth}>{\raggedright\arraybackslash}p{0.22\textwidth}>{\raggedright\arraybackslash}p{0.25\textwidth}Y}
\toprule
Control plane & Core issue & Main status and time scale & Typical failure mode \\
\midrule
Topology object layer & What experts are there, how big are they and how are they grouped or shared & Weight structure; slow variables in architecture design/training period & Knowledge redundancy, too coarse granularity, too heavy sharing paths \\
Routing decision-making layer & Which experts should be selected for the current token, how many to activate & affinity, Top-$k$, candidate constraints; token-by-token fast variables & misrouting, discrete optimization is difficult, semantics are topologically constrained \\
Balance control layer & Whether the usage distribution of a set of tokens meets training and system goals & expert/device/node load; sequence to runtime multi-scale feedback & collapse, over-uniformity, straggler, communication hotspot \\
Expert Parallel execution layer & Where, with what communication pattern and kernels, the selected expert is executed & placement, dispatch/combine, scheduling; step/runtime variables & exposed All-to-All time, small GEMMs, P99 latency, GPU-memory hotspots \\
\bottomrule
\end{tabularx}
\end{table}

There are two types of reverse dependencies between the four layers that cannot be ignored. First, the physical topology will constrain semantic routing: in order to reduce fan-out, device-limited routing actively reduces the candidate set visible to a certain token in $\mathcal{E}$. Second, Balance does not just tune the Router: runtime replication or expert placement can improve the physical load without changing $\mathcal{K}_t$. Therefore, "load balancing" cannot only be understood as an auxiliary loss, and "Expert Parallel" is not a passive implementation that is intervened after the model structure is determined.

\section{Eight Evolutionary Milestones: Criteria, Mainline, and Branches}

The eight milestones are neither a fixed taxonomy copied from one survey nor a mechanical division by year. They are an analytical summary based on \term{migration of the dominant bottleneck}. A change qualifies as a milestone only if it satisfies three criteria. First, it introduces an independently adjustable design variable, such as Top-$k$, expert granularity, a shared path, or a dynamic active budget. Second, it moves the system's principal bottleneck, for example from compute growing linearly with the number of experts, to discrete routing and load imbalance, and then to All-to-All communication, small GEMMs, or weight I/O. Third, later architectures inherit the change, so it is more than a one-off implementation detail.

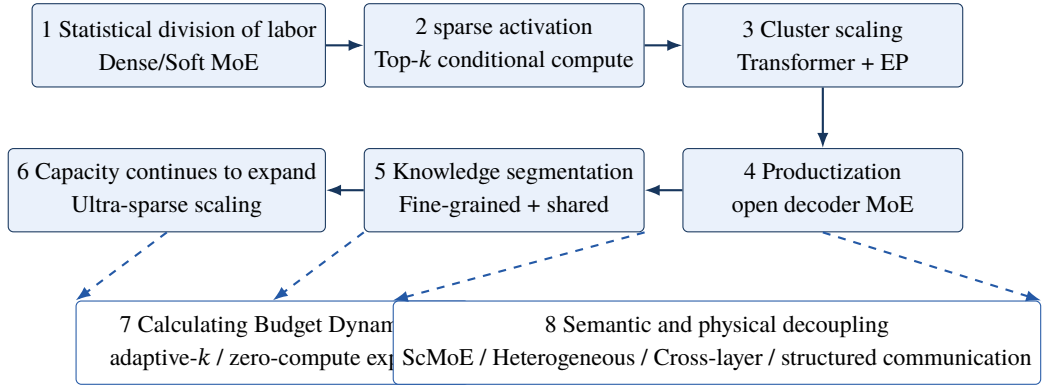
\begin{figure}[htbp]
\centering
\begin{tikzpicture}[
  node distance=5mm and 5mm,
  stage/.style={draw=DeepBlue,rounded corners=2pt,fill=SoftBlue,minimum width=37mm,minimum height=11mm,align=center,font=\small},
  branch/.style={draw=Accent,rounded corners=2pt,fill=white,minimum width=52mm,minimum height=11mm,align=center,font=\small},
  arr/.style={-{Latex[length=2mm]},thick,draw=DeepBlue},
  darr/.style={-{Latex[length=2mm]},thick,dashed,draw=Accent}
]
\node[stage] (s1) {1 Statistical division of labor\\Dense/Soft MoE};
\node[stage,right=of s1] (s2) {2 sparse activation\\Top-$k$ conditional compute};
\node[stage,right=of s2] (s3) {3 Cluster scaling\\Transformer + EP};
\node[stage,below=8mm of s3] (s4) {4 Productization\\open decoder MoE};
\node[stage,left=of s4] (s5) {5 Knowledge segmentation\\Fine-grained + shared};
\node[stage,left=of s5] (s6) {6 Capacity continues to expand\\Ultra-sparse scaling};
\node[branch,below=9mm of s6,xshift=14mm] (s7) {7 Calculating Budget Dynamics\\adaptive-$k$ / zero-compute experts};
\node[branch,below=9mm of s4,xshift=-14mm] (s8) {8 Semantic and physical decoupling\\ScMoE / Heterogeneous / Cross-layer / structured communication};
\draw[arr] (s1)--(s2);
\draw[arr] (s2)--(s3);
\draw[arr] (s3)--(s4);
\draw[arr] (s4)--(s5);
\draw[arr] (s5)--(s6);
\draw[darr] (s5.south west)--(s7.north);
\draw[darr] (s6.south)--(s7.north west);
\draw[darr] (s5.south east)--(s8.north west);
\draw[darr] (s4.south)--(s8.north east);
\end{tikzpicture}
\caption{The relationship between eight evolution nodes. Solid lines represent primary inheritance on the historical trunk, and dashed lines represent stackable orthogonal branches; node numbers do not represent strict generational replacement.}
\label{en:fig:evolution-dag}
\end{figure}

According to this standard, node 1--6 forms a clearer historical backbone: statistical division of labor $\rightarrow$ sparse activation $\rightarrow$ Transformer scale $\rightarrow$ decoder-only productization $\rightarrow$ fine-grained knowledge organization $\rightarrow$ ultra-sparse capacity expansion. Node 7 and node 8 are not simply new "generations" that follow node 6: Node 7 relaxes the assumption that "each token has a fixed amount of calculation"; node 8 relaxes the assumption that "semantic routing, intra-layer expert topology and physical communication must be bound". Both can be combined with the expert structure of node 5 or 6. Figure \ref{en:fig:evolution-dag} shows this inheritance relationship.

\begin{table}[htbp]
\centering
\caption{Division basis of eight nodes and bottleneck migration}
\label{en:tab:evolution-nodes}
\small
\begin{tabularx}{\textwidth}{p{0.055\textwidth}p{0.18\textwidth}p{0.25\textwidth}Y}
\toprule
Node & New design variable & Dominant problem solved & New bottleneck transferred out \\
\midrule
1 & gating and expert specialization & It is difficult for a single model to model the input space partition & All experts calculate, and the capacity is bound to FLOPs \\
2 & sparse Top-$k$ and capacity & total parameters decoupled from active FLOPs & discrete routing, collapse, token drop \\
3 & EP, All-to-All, stabilization loss & Cluster-level training of Transformer MoE & Communication, straggler, numerical and load stability \\
4 & decoder-only training and deployment closed loop & proves that classic MoE can enter the open model ecosystem & big expert knowledge redundancy, coarse combination granularity \\
5 & fine-grained, shared, device limit & Public knowledge duplication and insufficient expert combination & fan-out, small GEMM, shared path fixed cost \\
6 & Larger $N$, tiny experts, retrieval routing & Continue to expand total capacity under fixed active budget & Weighted I/O, retrieval, low token/expert \\
7 & adaptive compute / zero slots & token has different difficulty but uses fixed Top-$k$ & active budget control, tail delay, training stability \\
8 & shortcut, heterogeneous/cross-layer topology, structured communication & Semantic routing is bound by device topology and communication & Scheduling complexity, cache consistency, large-scale verification \\
\bottomrule
\end{tabularx}
\end{table}

\subsection{Dense/Soft MoE: Statistical division of labor rather than computational sparsity}

Jacobs et al. proposed the basic form \cite{en:jacobs1991adaptive} of gating network and local experts in 1991. What this node establishes is \term{ statistical division of labor }: gate generates mixed weights according to the input, different sub-networks fit different areas of the input space, and the training goals can promote specialization. Since all experts usually participate in weighting, routing is continuously differentiable, requiring neither token capacity nor sparse dispatch in the All-to-All sense.

\textbf{Why is this a separate milestone?} It established three roles retained by later MoE systems: a router or gate, a set of experts, and a weighted combination operator. \textbf{Why is it not yet a modern sparse MoE?} Increasing the number of experts still increases compute approximately linearly, so model capacity and FLOPs per token remain coupled. The next milestone preserves statistical specialization while executing only a small subset of experts, enabling conditional scaling but introducing optimization and systems problems associated with discrete selection.

\subsection{Sparse conditional computation: Top-k establishes capacity leverage}

Shazeer et al. applied noisy Top-$k$ routing to very large sparse networks\cite{en:shazeer2017outrageously}, so that only a few experts execute for each input. If the total number of experts is $N$ and each input activates only $k\ll N$ experts, total capacity can grow with $N$ while the dominant expert FLOPs scale approximately with $k$. This establishes the central capacity lever of modern MoE: \term{parameter capacity is decoupled from per-token compute}.

This decoupling is not free. Top-$k$ causes unselected experts to have no gradient from this token; popular experts will overflow capacity, and unpopular experts may not be trained for a long time; dispatch/combine must also be added for cross-device execution. Therefore, noisy routing, importance/load auxiliary loss, capacity factor and token drop are not peripheral techniques, but supporting mechanisms induced by sparse execution itself. Node 3 does not change this basic algorithm, but answers: how to make MoE stable and executable when it is repeatedly embedded in Transformer layers and scaled to thousands of devices.

\subsection{Transformer MoE and Expert Parallel}

GShard systematically embeds Top-2 MoE FFN into Transformer and relies on automatic sharding to train \textbf{ on \textbf{2048 TPU} multi-language model \cite{en:lepikhin2020gshard} with more than 600B}. The classic execution chain is thus fixed as
\begin{center}
\textsf{Attention $\rightarrow$ Router $\rightarrow$ Dispatch All-to-All $\rightarrow$ Expert FFN $\rightarrow$ Combine All-to-All}.
\end{center}
The new variable of this node is not "more experts", but \term{expert parallelism and cluster execution semantics }: how tokens are rearranged across devices, how each expert is batch-processed, how overflows are handled, and how communication and calculation are synchronized. Switch Transformer further adopts Top-1, trading lower communication and simpler execution for expert combination capabilities and routing fault tolerance reduction \cite{en:fedus2021switch}. ST-MoE elevates stability to a first-class design goal and introduces Router z-loss constraint logits numerical scale \cite{en:zoph2022stmoe}. It needs to be distinguished: z-loss controls numerical stability, and balance loss controls usage distribution. The two are not the same mechanism.

During the same period, BASE Layers wrote the training route as a strictly balanced linear allocation problem \cite{en:lewis2021base}, while Expert Choice allowed experts to choose fixed capacity tokens \cite{en:zhou2022expertchoice}. They can directly guarantee equilibrium, but batch-level joint allocation and expert-side capacity are not as natural as token-choice for online autoregressive decoding, so they did not replace the Top-$k$ backbone of decoder-only LLM. The legacy of node 3 is the complete system contract of "Router + EP + capacity/balance + All-to-All"; nodes 4 and 5 both follow it, but shift the focus from "can large-scale training" to model quality, deployment availability and expert internal organization.

\subsection{Open-Weight Coarse-Grained MoE}

Mixtral uses \textbf{8$\times$7B parameters with Top-2 routing per token}\cite{en:jiang2024mixtral}. It did not introduce a new router family and therefore is not a milestone purely in terms of algorithmic novelty. Its importance is different: it demonstrated that \term{a conventional coarse-grained MoE can support pretraining, instruction tuning, inference deployment, and community reproduction} in an open-weight decoder-only LLM.

Why is this a \textbf{separate milestone}? Node 3 demonstrates scalable training on very large clusters; node 4 establishes an end-to-end path from pretraining and instruction tuning to practical deployment for a general-purpose decoder model. It inherits Top-$k$, isomorphic experts, and EP without changing the basic execution chain. Once the system can run reliably, the main bottleneck shifts to knowledge organization: a small number of large experts tend to relearn common capabilities, while the Router can compose only coarse knowledge blocks. Node 5 addresses this bottleneck directly.

\subsection{Shared + fine-grained experts}

DeepSeekMoE makes two key changes to coarse-grained experts\cite{en:dai2024deepseekmoe}. First, it splits a large FFN into multiple smaller experts, allowing the Router to compose more knowledge units under the same active budget. Second, shared-expert isolation moves knowledge needed by all tokens onto an always-on path, reducing redundancy among routed experts. The former increases \term{composition resolution}, while the latter explicitly separates common and conditional capabilities. DeepSeek-V2 scales this structure to \textbf{236B total parameters and 21B active parameters}, and adds device-limited routing: each token's candidate experts are restricted to a small number of devices, thereby controlling cross-device fan-out\cite{en:deepseekv2}.

Shared or dense paths can also provide computational windows that hide communications. DeepSeek-V2 combines shared expert computation with EP communication overlap; Snowflake Arctic's dense residual path also reflects the similar algorithm -- system collaboration \cite{en:snowflakearctic}. But \term{shared expert is not a necessary condition for fine-grained MoE }. Qwen3-235B-A22B uses \textbf{128 fine-grained experts and Top-8}, but removes the shared expert and relies on global-batch balance to allow commonly used capabilities to naturally form \cite{en:qwen3} among routed experts.

Therefore, node 5 actually contains two divisible axes: expert granularity and shared path. They are often combined, but there is no logical binding. Node 5 inherits the EP system of node 3, but increases the number of experts, Top-$k$, and device fan-out at the same time, making small GEMM, routing fragmentation, and communication topology new bottlenecks. Node 6 continues to expand along the lines of "more and smaller experts"; Node 8 attempts to change the physical implementation of fan-out and communication.

\subsection{Ultra-sparse scaling}

The question at this stage is: \emph{with active parameters fixed, does increasing the total number of experts continue to reduce loss?} PEER uses product-key retrieval to select a small subset from \textbf{more than one million tiny experts}\cite{en:he2024peer}. Kimi K2 scales an engineering-oriented ultra-sparse design to \textbf{1.04T total parameters, 32.6B active parameters, 384 routed experts, and Top-8 routing}, and reports scaling gains from higher sparsity at a fixed active budget\cite{en:kimiK2}. GPT-OSS-120B uses \textbf{128 experts with Top-4 routing} and a relatively small active footprint, illustrating an ultra-sparse design intended for local deployment\cite{en:openaiGptOss}.

\textbf{Why separate ultra-sparse scaling from fine-grained experts?} Fine-grained experts primarily reduce knowledge redundancy and increase composition resolution. Ultra-sparse scaling instead asks whether a larger candidate expert set continues to improve loss when active parameters and training FLOPs are fixed. Both increase expert count, but they optimize different objectives.

\term{Sparsity cannot increase without bound.} As $N$ grows, router scoring, expert-weight I/O, All-to-All metadata, and small-GEMM fragmentation all increase. When each expert receives too few tokens per step, \textbf{theoretical FLOP savings no longer translate into wall-clock gains}. Ultra-sparse scaling therefore depends more heavily on fine-grained organization and routing control rather than replacing them. The next two milestones relax different constraints of the fixed Top-$k$ mainline.

\subsection{Dynamic compute: from fixed Top-k to token-dependent budget}

Fixed Top-$k$ implies that every token receives the same number of expert FLOPs even when tokens differ in semantic difficulty and required capacity. LongCat-Flash introduces zero-computation experts: the router still selects a fixed number of slots, but some slots are identities, so actual FFN compute varies by token\cite{en:longcatFlash}. Adaptive-$k$, threshold routing, and zero-compute slots share the objective of \term{decoupling which experts are selected from how much computation is allocated}.

\textbf{Relationship to ultra-sparse scaling.} Ultra-sparse scaling expands the candidate capacity $N$ while usually keeping $k$ fixed; dynamic computation changes the actual $k_t$, or the effective FFN compute, for each token. The two approaches can be combined: the candidate pool remains large, while easy tokens receive less compute and difficult tokens receive more. The control objective consequently expands from expert load alone to the mean active budget, budget variance, per-expert load, and P99 latency. Constraining average training FLOPs does not guarantee tail latency because difficult tokens may cluster within a micro-batch or request window.

The remaining issue of dynamic computing is closed-loop control: if the Router determines both the semantic expert and the computing budget, the balance bias affects both specialization and throughput; if only the long-term average is constrained, short-term bursts may occur. Therefore, adaptive bias/PID feedback, budget loss and runtime admission control are supporting issues and cannot be solved by simply modifying Top-$k$.

\subsection{Semantic routing decoupled from physical execution}

Node 1--7 mainly changes Router's optional expert or active budget; node 8 changes the lower-level assumption: \term{ semantically selecting an expert does not mean that it must be executed immediately along the fixed intra-layer All-to-All path }. This node is not a single model family, but four frontier branches with a common goal.

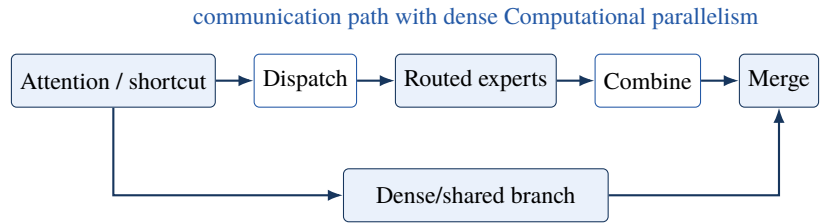
\begin{figure}[htbp]
\centering
\begin{tikzpicture}[
  node distance=5mm and 5mm,
  box/.style={draw=DeepBlue,rounded corners=2pt,fill=SoftBlue,minimum height=7mm,align=center,font=\small},
  comm/.style={draw=Accent,rounded corners=2pt,fill=white,minimum height=7mm,align=center,font=\small},
  arr/.style={-{Latex[length=2mm]},thick,draw=DeepBlue}
]
\node[box] (attn) {Attention / shortcut};
\node[comm,right=of attn] (dispatch) {Dispatch};
\node[box,right=of dispatch] (expert) {Routed experts};
\node[comm,right=of expert] (combine) {Combine};
\node[box,right=of combine] (merge) {Merge};
\node[box,below=8mm of expert,minimum width=35mm] (dense) {Dense/shared branch};
\draw[arr] (attn)--(dispatch);
\draw[arr] (dispatch)--(expert);
\draw[arr] (expert)--(combine);
\draw[arr] (combine)--(merge);
\draw[arr] (attn.south) |- (dense.west);
\draw[arr] (dense.east) -| (merge.south);
\node[font=\small\color{Accent},above=2mm of expert] {communication path with dense Computational parallelism};
\end{tikzpicture}
\caption{At its core, ScMoE is about rearranging execution dependencies to expand the communication hidden window; it is not a new routing algorithm.}
\label{en:fig:scmoe}
\end{figure}

\textbf{Execution-dependency rearrangement.} ScMoE does not change which experts a token selects; it changes the dependency graph. Its cross-layer shortcut lets routed experts consume an intermediate representation from the preceding layer while the current dense/shared path overlaps the dispatch/combine communication window\cite{en:cai2024scmoe}. It reduces exposed communication but does not directly solve routing skew. Figure~\ref{en:fig:scmoe} gives an abstract comparison.

LongCat-Flash combines ScMoE with token-dimension chunking and Single Batch Overlap; the official LongCat-2.0 report further adopts fully parallel dense/MoE execution on each core of a dedicated accelerator\cite{en:longcat2}. These developments reduce exposed communication, \textbf{but they do not automatically eliminate severe routing skew, heterogeneous expert runtimes, or GPU-memory pressure}.

\textbf{Heterogeneous experts.} MoHGE uses two-level routing: it first selects an expert group and then experts of different sizes within that group, together with group-wise and intra-group balancing constraints\cite{en:mohge2026}. Equal expert width becomes a design variable rather than a default constraint, allowing different tokens to receive different forms of compute. The cost is greater complexity in heterogeneous GEMMs, capacity planning, and load metrics.

\textbf{Cross-layer sharing.} GMoE designates some experts as cross-layer Global Experts while retaining layer-local experts, reducing redundant capabilities across layers\cite{en:gmoe2026}. This differs from a conventional shared expert in scope: the former is reused across layers, whereas the latter is typically an always-on path within one layer. Cross-layer sharing introduces additional caching, scheduling, and inter-layer interference concerns.

\textbf{Structured communication.} Multi-Head LatentMoE and Head Parallel seek a topology whose communication volume is $O(1)$ with respect to the number of activated experts $k$, yielding more predictable traffic\cite{en:cui2026latentmoe}. Whereas ScMoE hides dynamic communication, this route attempts to reduce or structure the communication itself. In the absence of public validation at trillion-parameter and ten-thousand-accelerator scale, \term{it should be treated as a research frontier, not as a mainstream replacement for Top-$k$ expert parallelism}.

The four branches respectively optimize temporal overlap, computation structure, parameter reuse, and data movement. They are complementary and can be combined.

\section{Expert topology: object layer}

In the four-layer framework introduced in Chapter 2, Topology is the object layer. It defines the expert set $\mathcal{E}$, the granularity of each knowledge unit, the sharing scope, and computational heterogeneity before Routing makes token-level selections. The central question is therefore not simply how many experts exist, but \term{what kinds of computational units the router can compose}.

It is not sufficient to understand the evolution of MoE only as the growth of expert count. Table \ref{en:tab:topology} shows that what really changes is the granularity, sharing scope and computing heterogeneity of knowledge units.

\begin{table}[htbp]
\centering
\caption{The main evolution route of Expert topology}
\label{en:tab:topology}
\small
\begin{tabularx}{\textwidth}{p{0.21\textwidth}p{0.23\textwidth}Y Y}
\toprule
Topology & Representative work & Main benefit & New bottleneck \\
\midrule
A small number of isomorphic experts & GShard, Switch, Mixtral & Realize conditional scaling & Knowledge duplication, limited combinability \\
Shared/dense + routed & DeepSeekMoE, Arctic & Public knowledge isolation, can create overlap window & Fixed FLOPs increase, public path may be overweight \\
Fine-grained experts & DeepSeek-V2/V3, Qwen3 & Fine-grained experts combination, higher total/active ratio & Routing, communication and GEMM updated \\
Ultra-sparse experts & PEER, Kimi K2, gpt-oss & Continue to expand capacity under fixed FLOPs & Retrieval, weighted I/O, low token/expert \\
Heterogeneous/dynamic & LongCat, MoHGE & Token-level dynamic computing budget & budget control coupled with physical equalization \\
Cross-layer global/local & GMoE & Reduce cross-layer expert redundancy & Scheduling, caching and inter-layer interference need to be verified \\
Latent/head structured & Multi-Head LatentMoE & Communication deterministic, decoupled from $k$ & Insufficient evidence of large-scale quality \\
\bottomrule
\end{tabularx}
\end{table}

Another dimension that is often overlooked is the initialization method. Sparse Upcycling copies the existing dense FFN into multiple experts, and then adds a new Router to continue training \cite{en:komatsuzaki2022upcycling}. It reuses dense checkpoints, but brings expert symmetry. It requires routing noise, data differences and subsequent training to differentiate the experts. Further Expert Upcycling is extended from the existing $E$-expert MoE to $mE$ experts, while keeping Top-$k$ and active cost unchanged \cite{en:expertUpcycling2026}. Upcycling is not a new Router family, but it will change the formation path and training cost of specialization.

\section{Routing: decision-making layer}

In the closed loop of Figure~\ref{en:fig:fourplanes}, Routing receives the expert set defined by Topology and the candidate constraints imposed by the system, and generates $\mathcal{K}_t$ for each token. It optimizes \term{local semantic matching}; it does not directly guarantee a reasonable aggregate load over a sequence, batch, or device. The latter belongs to the Balance control plane discussed in Section~6.

\begin{table}[htbp]
\centering
\caption{Main Router families and their applicable boundaries}
\label{en:tab:router}
\small
\begin{tabularx}{\textwidth}{p{0.24\textwidth}p{0.19\textwidth}Y Y}
\toprule
Router family & Decision-making unit & Balance characteristics & decoder-only LLM Applicability \\
\midrule
Noisy Top-$k$ token-choice & token selection experts & requires loss, bias or runtime balance & current trunk, online naturally \\
Top-1 Switch & token choose 1 expert & Simple communication, more sensitive capacity & Expandable, but limited combination ability \\
Balanced assignment & batch joint assignment & Strictly equal during training & Global solution is complex and unnatural online \\
Expert-choice & expert select tokens & expert capacity is naturally fixed & encoder is more natural, decoder capacity is difficult to process \\
Soft/parameter merging & Continuous mixing & Differentiable, non-dispersion Top-$k$ & Large-scale EP efficiency not yet proven \\
Bias-controlled Top-$k$ & affinity + non-gradient bias & does not write the balanced gradient into the LM parameter & DeepSeek-V3 and other models use \\
Zero experts/adaptive-$k$ & token The amount of calculation is variable & The average budget and tail need to be constrained & The important direction of dynamic calculation \\
\bottomrule
\end{tabularx}
\end{table}

The fact that Token-choice has become mainstream does not mean that it is optimal in terms of optimization, but that it is the most consistent with the autoregressive online constraints: the routing of the current token only relies on the current state, and there is no need to wait for future tokens or solve the global distribution of the entire batch. Continuous routing schemes such as Soft MoE and Lory avoid discrete Top-$k$, but the system advantages have not yet been proven on ultra-large-scale decoder and low-latency EP\cite{en:puigcerver2023softmoe,en:zhong2024lory}.

Routing freedom is also constrained by network topology. Device-limited or node-limited routing limits candidate experts to a small number of devices/nodes to reduce single-token communication fan-out. The trade-off is that the semantically optimal expert may be excluded by topological constraints, so this type of design is inherently an explicit trade-off between quality and communication.

\section{Balance: control layer}

Balance does not redefine experts, nor does it independently generate semantic routes for tokens. It observes the $\{n_i\}$ and equipment, nodes and communication loads accumulated from many routing decisions, and then feeds back to the decision-making and execution layers through loss, bias, capacity or runtime placement. The core problem is: while \term{ allows uneven semantic specialization, how to avoid training collapse and physical straggler}.

\subsection{Sequence/batch-wise expert balance}

The classic auxiliary loss can be written as
\begin{equation}
  \mathcal{L}_{\mathrm{bal}}=\alpha N\sum_{i=1}^{N} f_i P_i,
  \label{en:eq:balance}
\end{equation}
Among them, $f_i$ is the token fraction actually assigned to expert $i$ in the statistical window, $P_i$ is the average routing probability, and $\alpha$ controls the equalization intensity. When all experts use approximately uniform values, the formula ~\eqref{en:eq:balance} is smaller. This loss is responsible for both preventing expert collapse and avoiding EP straggler, but they are not equivalent: the model may require uneven specialization, but the hardware hopes to have an even load. When $\alpha$ is too large, the hardware target directly interferes with the language modeling gradient.

\subsection{Device/node/communication-aware balance}

DeepSeek-V2 splits the balance into expert-level, device-level and communication balance, and controls fan-out\cite{en:deepseekv2} through device-limited routing. The key to this change is not to add a few more losses, but to acknowledge that \term{ logical expert balance is not equal to physical device or network balance }: If there are multiple experts on a device, as long as the total tokens between devices are close, it is not necessarily necessary that all experts are completely equal.

\subsection{Auxiliary-Loss-Free Load Balancing}

The Auxiliary-Loss-Free (ALF) strategy adds non-gradient routing bias $b_i$ to each expert:
\begin{equation}
  \mathcal{K}_t=\operatorname{TopK}_i(s_{i,t}+b_i),\qquad
  b_i\leftarrow b_i+\eta\,\operatorname{sign}(\bar n-n_i),
  \label{en:eq:alf}
\end{equation}
Where $\eta$ is the bias update speed. The bias affects the Top-$k$ selection, but does not participate in the LM gradient as an affinity weight, thereby reducing the direct interference of the equilibrium target on the model parameter update direction \cite{en:wang2024auxfree}. DeepSeek-V3 uses batch-wise ALF as the main balancing mechanism, while retaining a very weak sequence-wise auxiliary loss to prevent extreme single-sequence imbalance \cite{en:deepseekv3}. Therefore, \term{ "DeepSeek-V3 has no balance loss at all" is not accurate }.

Global-batch statistics are more stable than single micro-batch, and also allow some experts to carry more specific domain tokens on local batches; the cost is that load needs to be aggregated across DP/EP ranges, and expanding the statistical range may introduce additional communication.

\subsection{Runtime expert placement}

Balance on the training distribution cannot guarantee true inference traffic. Domain shift, SFT/RL and off balance updates all change expert popularity. DeepSeek-V3 uses redundant expert deployment, replicates hotspot experts and adjusts placement\cite{en:deepseekv3} at runtime. This trend re-partitions the goals: Routers can preserve semantically non-uniform specialization, and runtimes avoid device stragglers by copying, migrating, or relocating them.

\begin{figure}[htbp]
\centering
\begin{tikzpicture}[
  stage/.style={draw=DeepBlue,fill=SoftBlue,rounded corners=2pt,minimum width=30mm,minimum height=10mm,align=center,font=\small},
  arr/.style={-{Latex[length=2mm]},thick,draw=Accent}
]
\node[stage] (s1) {Sequence/Batch\\expert balance};
\node[stage,right=7mm of s1] (s2) {Device/Node/\\Communication};
\node[stage,right=7mm of s2] (s3) {Global-batch\\ALF bias};
\node[stage,right=7mm of s3] (s4) {Runtime replication\\and placement};
\draw[arr] (s1)--(s2);
\draw[arr] (s2)--(s3);
\draw[arr] (s3)--(s4);
\end{tikzpicture}
\caption{The evolution of Balance is not to simply weaken the constraints, but to gradually separate the physical execution goals from the LM gradient.}
\label{en:fig:balance}
\end{figure}
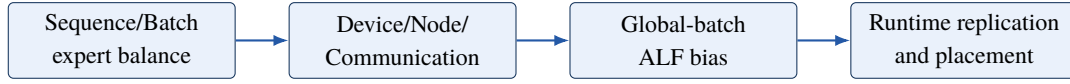

\section{Expert Parallel: Execution Layer}

EP consumes the token--expert assignment generated by Routing, and combines it with placement $\pi$ to convert the logical expert into dispatch, local GEMM and combine on the device. It usually does not change "who is semantically selected", but determines the wall-clock cost of this selection; when the system cost is too high, the first three layers will be reversely modified through device-limited routing, placement or architectural reconstruction.

The algorithmic benefits of MoE only hold true if the sparse execution is efficient. The capacity-based system reserves fixed token slots for each expert, and overflow tokens are discarded or used as residual, which can easily cause inconsistency between training and inference. MegaBlocks uses block-sparse kernels to implement dropless MoE, so that dynamic token shapes no longer require token dropping\cite{en:gale2022megablocks}. Systems such as Tutel dynamically select parallel and pipeline strategies to adapt to different cluster sizes.\cite{en:hwang2022tutel}.

As experts become thinner and Top-$k$ becomes larger, communication gradually becomes a first-class architectural constraint, driving three processing methods:
\begin{enumerate}
\item \term{ Limited communication topology }: device/node-limited routing reduces fan-out;
\item \term{ Hidden communication }: shared computation overlap, DualPipe, ScMoE and token chunking expand the overlap window;
\item \term{ Reconstruct communication }: Head Parallel and other solutions turn dynamic traffic into a more deterministic communication mode.
\end{enumerate}

Here \term{ must distinguish between "total All-to-All time" and "exposed communication time" }. If the communication is completely covered by the dense path, the marginal benefit of continuing to optimize the link bandwidth on the end-to-end step time will decrease; conversely, even if the total communication volume remains unchanged, as long as the critical path is shortened, the throughput may be significantly improved. Therefore, the training report should at least give the \textbf{ traffic volume, overlap ratio, exposed dispatch/combine, MFU and step-time P99}.

\section{The Modern Mainline and Frontier Branches}

As of 2026, the most common combinations of large-scale decoder MoE can be summarized as:
\begin{center}
\fcolorbox{DeepBlue}{SoftGray}{\parbox{0.88\textwidth}{\centering
Token-choice Top-$k$ + fine-grained experts + optional shared path + global/batch-wise balance or ALF + topology-limited routing + dropless kernels/communication overlap + runtime expert placement}}
\end{center}

DeepSeek-V3, Qwen3 and Kimi K2 differ in shared expert, balance mechanism and EP size, but they are all located on this backbone. At the same time, the following branches have not yet reached a unified conclusion:
\begin{itemize}
\item \term{MoE Attention}: JetMoE sparses Attention and FFN at the same time, further reducing active FLOPs, but KV/cache and routing are more complex \cite{en:shen2024jetmoe};
\item \term{Hybrid mixer + MoE}: Jamba alternates between Transformer and Mamba layers, indicating that MoE is a parameter expansion method and does not bind the full Attention backbone\cite{en:lieber2024jamba};
\item \term{Million experts}: High theoretical composability, but weighted I/O, small GEMM and retrieval overhead still limit industrial deployment;
\item \term{Heterogeneous experts}: Choose different capacities according to token difficulty, with natural expression, but physical placement and tail delay are more difficult;
\item \term{Cross-layer global experts}: The goal is to eliminate inter-layer duplication capabilities, but there is still a lack of evidence for very large-scale training;
\item \term{Structured communication}: Head Parallel may reduce the pain point of dynamic EP, but it is still in the early verification stage.
\end{itemize}

Table \ref{en:tab:models} compares representative models based on architecture selection rather than benchmark ranking.

\begin{table}[htbp]
\centering
\caption{Architectural location of representative LLM MoE; active parameter size varies by report}
\label{en:tab:models}
\scriptsize
\begin{tabularx}{\textwidth}{p{0.18\textwidth}C{15mm}C{25mm}C{22mm}Y}
\toprule
Model & Year & Total/Active & Experts/Top-$k$ & Architectural significance \\
\midrule
GShard & 2020 & $>$600B/Not uniformly disclosed & Top-2 & Transformer MoE with automatic sharding \\
Switch & 2021 & up to 1T+ & Top-1 & Exchange throughput and scalability with simple routing \\
Mixtral 8$\times$7B & 2024 & 46.7B/12.9B & 8/2 & Open coarse-grained MoE Benchmark \\
DeepSeek-V2 & 2024 & 236B/21B & 160/6 & fine-grained, shared, device-limited \\
DeepSeek-V3 & 2024 & 671B/37B & 256/8 & ALF, weak seq loss, node-limited \\
Qwen3-235B-A22B & 2025 & 235B/22B & 128/8 & fine-grained, no shared, global balance \\
Kimi K2 & 2025 & 1.04T/32.6B & 384/8 & ultra-sparse scaling \\
gpt-oss-120b & 2025 & 116.8B/5.1B & 128/4 & Small active footprint and deployment orientation \\
LongCat-Flash & 2025 & 560B/18.6--31.3B & real + zero & Dynamic Compute Budget with ScMoE \\
LongCat-2.0 & 2026 & 1.6T/33--56B & dynamic & dense/MoE per-core parallel \\
\bottomrule
\end{tabularx}
\end{table}

\section{Experimental Design for Foundation-Model Pretraining}

\term{ cross-model benchmark cannot identify architectural contributions. } To determine whether fine-grained, shared, ALF or ScMoE is effective, you should at least fix the \textbf{ training token, data ratio, active parameters and optimizer } at the same time, and report \textbf{training-FLOPs-matched and wall-clock-matched} result.

\begin{table}[htbp]
\centering
\caption{Proposed equal-budget MoE ablation matrix}
\label{en:tab:experiments}
\small
\begin{tabularx}{\textwidth}{p{0.20\textwidth}p{0.39\textwidth}Y}
\toprule
Experiment dimension & Recommendation setting & Core indicator \\
\midrule
Expert granularity & Fixed total/active params and tokens, scan expert count, width, Top-$k$ & val loss, downstream, token/expert, GEMM efficiency \\
Shared vs emergent & shared; no shared+global loss; no shared+ALF & expert similarity, domain-routing MI, active FLOPs \\
Balance scope & sequence aux, global aux, ALF, ALF+weak seq & max/mean load, CV/Gini, entropy, specialization \\
Execution topology & serial EP, shared overlap, ScMoE, runtime EPLB & exposed communication, MFU, step time, TPOT, P99 \\
Dynamic compute & fixed Top-$k$, threshold, zero experts & active-param mean/variance, bucketing loss, throughput and tail delay \\
Upcycling & scratch, dense-copy, cluster/utility-aware copy & Symmetry breaking speed, loss/token, GPU hours \\
\bottomrule
\end{tabularx}
\end{table}

For code pre-training scenarios, it is recommended to additionally bucket statistics by language, repo/file/function granularity, code and natural language token routing. Key observations include: expert co-activation in different languages, whether cross-file reference tokens are concentrated in a small number of experts, whether code and code-related documents share experts, and whether long repo-level sequences cause sequence-wise load peaks. Only in this way can we distinguish between "experts forming beneficial specialization" and "data matching causing accidental traffic skew".

The system side should at least record at the same time:
\begin{itemize}
\item max/mean, CV and Gini at the three levels of expert, device and node;
Load under the three statistical windows of \item sequence, micro-batch, and global batch;
\item dispatch/combine exposed time, not just the total All-to-All time;
  \item token drop, capacity overflow, redundant expert hit rate;
\item training FLOPs, active parameters, wall-clock and cluster network topology.
\end{itemize}

\section{Open Questions}

\subsection{What is the correct unit of analysis for Specialization?}

"Mathematical expert" or "code expert" are intuitive narratives, but actual routing is often represented by a combination of token morphology, syntax, location, language, and hierarchy. A single expert may not be a stable semantic unit, and a cross-layer expert group or co-activation pattern may be more interpretable. In the future, routing mutual information, expert representation similarity and intervention-based causal test should be combined instead of just showing token word cloud.

\subsection{Where Is the Systems Optimum at Higher Sparsity?}

Increasing total experts under fixed active FLOPs may reduce model loss, but system benefits are limited by network bandwidth, expert weight I/O, batch size and kernel arithmetic intensity. The optimal sparsity is not a pure model constant, but the model -- hardware joint scaling law. The EP size, network topology, and number of tokens per expert must be given when reporting this conclusion.

\subsection{Can Load Balancing Move Primarily to Runtime?}

ALF reduces the gradient interference of the auxiliary loss, but still changes the Top-$k$ selection. Runtime replication and placement can allow more freedom in semantic routing, but will bring parameter copy memory, migration costs and cache consistency issues. The optimal boundary between weak constraints in training and strong equilibrium at runtime remains undetermined.

\subsection{How Dynamic Compute Meets Tail Latency SLAs}

Zero experts or adaptive-$k$ can reduce average FLOPs, but difficult tokens will create new stragglers if they are in the same micro-batch or request window. The model needs to control average budget, budget variance, and P99 simultaneously, rather than just reporting average active parameters.

\subsection{How Can Post-Training Preserve Router Stability?}

SFT/RL data will change the domain distribution and sequence shape, thereby changing the expert load. Whether to retain balance updates, routing replay, and freeze Routers during the training phase cannot be directly extrapolated from the pre-training settings. Routing drift and specialization retention should be measured separately in the pretrain, SFT and RL stages.

\section{Summary and Outlook}

The architectural history of \term{LLM MoE is not a single-line expansion of "more and more experts", but the co-evolution of five routes: }expert has changed from a small number of large modules to fine-grained, ultra-sparse and heterogeneous computing units; the topology has developed from layer-local routed experts to shared, cross-layer global/local and latent-head structures; Router has evolved from simple Top-$k$ gradually adds global statistics, bias control and dynamic computing budget; balance moves from sequence-wise auxiliary loss to device/communication-aware, ALF and runtime placement; execution develops from serial All-to-All to dropless kernels, topology-limited routing, ScMoE overlap and structured communication.

Therefore, there is no unique "next-generation structure" after ScMoE. A more credible direction is a combination of four: freer semantic routing, explicit average computing budget control, topology awareness or deterministic communication, and runtime expert placement. Architecture evaluation must also be expanded from a single validation loss to a joint analysis of quality, specialization, active budget, exposed communication, and tail latency.

For the next round of large-scale pretraining, the most informative experiment is not a direct comparison of final benchmark scores from two different recipes. Instead, construct a controlled matrix with \textbf{the same data, token budget, and active FLOPs}, then vary expert granularity, the shared path, balancing scope, and execution topology separately. \textbf{This separation is necessary to distinguish gains from parameter capacity, routing specialization, and systems throughput}, and to derive MoE design conclusions that transfer to new hardware.

\appendix
\section{Architecture Evolution Cheat Sheet}

\begin{longtable}{@{}p{0.12\textwidth}p{0.19\textwidth}p{0.28\textwidth}p{0.30\textwidth}@{}}
\caption{Bottleneck migration of MoE architecture evolution}\label{en:tab:timeline}\\
\toprule
Period & Representative work & Main problems solved & New bottlenecks transferred out \\
\midrule
\endfirsthead
\toprule
Period & Representative work & Main problems solved & New bottlenecks transferred out \\
\midrule
\endhead
1991--2016 & Adaptive Mixtures & Learning soft division of labor & Computation increases linearly with the number of experts \\
2017--2019 & Sparse MoE & Parameter capacity and single token FLOPs decoupling & collapse, capacity, All-to-All \\
2020--2022 & GShard, Switch, ST-MoE & Transformer Large-scale sparse training & Stability, token drop, routing quality \\
2023--2024 & Mixtral & Open decoder MoE availability & Coarse-grained and knowledge redundant \\
2024--2025 & DeepSeekMoE/V2, Qwen3 & Fine-grained combination and public knowledge processing & Communication fan-out, small GEMM, balance scope \\
2024--2025 & PEER, Kimi K2, gpt-oss & ultra-sparse capacity expansion & retrieval, I/O, low token/expert \\
2024--2026 & ALF, LongCat, ScMoE & Reduce gradient interference and hidden communication & Runtime dynamic load and tail delay \\
2025--2026 & MoHGE, GMoE, Head Parallel & Heterogeneous computing, cross-layer sharing, structured communication & Ultra-large-scale quality and system verification \\
\bottomrule
\end{longtable}

\clearpage
\pdfbookmark[0]{中文版}{bilingual.chinese}
\setcounter{section}{0}
\setcounter{subsection}{0}
\setcounter{subsubsection}{0}
\setcounter{equation}{0}
\setcounter{figure}{0}
\setcounter{table}{0}
\setcounter{footnote}{0}
\setcounter{tocdepth}{2}
\renewcommand{\abstractname}{摘要}
\renewcommand{\contentsname}{目录}
\renewcommand{\figurename}{图}
\renewcommand{\tablename}{表}
\renewcommand{\refname}{参考文献}
\renewcommand{\appendixname}{附录}
\BilingualUseChineseToc
\renewcommand*{\theHsection}{zh.\arabic{section}}
\renewcommand*{\theHsubsection}{zh.\arabic{section}.\arabic{subsection}}
\renewcommand*{\theHsubsubsection}{zh.\arabic{section}.\arabic{subsection}.\arabic{subsubsection}}
\renewcommand*{\theHequation}{zh.\arabic{equation}}
\renewcommand*{\theHfigure}{zh.\arabic{figure}}
\renewcommand*{\theHtable}{zh.\arabic{table}}
\gdef\BilingualAnchorPrefix{zh}
\fancyhead[L]{\small\color{gray}LLM MoE 架构演进}
\renewcommand{\reported}{\textbf{论文报告：}}
\renewcommand{\inference}{\textbf{本文判断：}}
\markboth{}{}
\begin{center}
  {\zihao{1}\bfseries\color{DeepBlue}
  LLM MoE 架构演进：从稀疏路由到动态计算与通信结构重构}\par
  \vspace{4pt}
  {\zihao{3}\color{DeepBlue} Expert Topology、Routing、Load Balance 与 Expert Parallel 的协同演化}\par
  \vspace{10pt}
  {\normalsize\textbf{Jiguo Li\footnote{本文在Codex协助下完成}}}\par
  \vspace{2pt}
  {\small\href{mailto:jiguolee@gmail.com}{jiguolee@gmail.com}}\par
  \vspace{6pt}
  {\small 2026 年 8 月}\par
\end{center}
\vspace{5pt}
\hrule

\begin{abstract}
Mixture-of-Experts（MoE）已经成为大语言模型在固定单 token 计算预算下扩展参数容量的核心架构之一。然而，若只按模型发布时间罗列 GShard、Switch、Mixtral、DeepSeekMoE、DeepSeek-V3 与新近动态 MoE，会掩盖真正推动结构变化的瓶颈迁移。本文综合算法、推理系统和高效架构三类 Survey，并回到代表性原始论文与官方技术报告，提出一个由五条耦合路线构成的统一框架：expert 粒度、expert 拓扑、路由自由度、负载均衡作用域和执行结构。基于该框架，本文把 MoE 演进归纳为八个演进节点；它们不是八代模型的线性年表，而是由六个主干节点与两个正交分支构成的依赖图。在历史时间轴之外，本文进一步用 Expert topology、Routing、Balance 和 Expert Parallel 四个控制面剖解任一代 MoE 的内部工作机制，分别回答“有哪些专家、每个 token 选谁、群体负载如何受控、选中的计算如何映射到设备”。最后给出适用于基座预训练的等预算实验设计、系统指标和仍待解决的研究问题。\textbf{核心结论是：现代 MoE 的竞争焦点已从“稀疏激活更多参数”转向“让语义路由、计算预算和物理执行解耦”。}
\end{abstract}

\noindent\textbf{关键词：}大语言模型；Mixture-of-Experts；稀疏路由；负载均衡；Expert Parallel；动态计算；ScMoE

\noteBox{\textbf{阅读提要：}第三章的八个节点描述历史上的瓶颈迁移；第四至七章则从 topology、routing、balance 和 Expert Parallel 四个控制面剖解单个 MoE 系统。前者是时间轴，后者是系统剖面，不能视为两套并列的“发展阶段”。}

\clearpage
\begingroup
\small
\setlength{\parskip}{0pt}
\linespread{1.02}\selectfont
\setcounter{tocdepth}{2}
\tableofcontents
\endgroup
\clearpage

\section{问题定义与分析范围}

经典 MoE 研究关注如何让 gating network 将样本分配给不同局部专家；LLM 时代的稀疏 MoE 则额外要求：总参数容量增长时，每个 token 的激活参数和计算量不能同比增长。现有三类综述分别提供互补视角：Cai 等人的 Survey 按算法、系统和应用建立全栈 taxonomy\cite{zh:cai2024survey}；Liu 等人重点拆解推理阶段的模型、系统与硬件优化\cite{zh:liu2024inference}; Zhu 等人则把 MoE 放回 sparse attention、SSM 和 hybrid architecture 的大图景中\cite{zh:zhu2025speed}。本文不重复逐篇枚举，而是追问：\textbf{每一次结构变化消除了什么瓶颈，又把瓶颈转移到哪里？}

讨论范围限定为 decoder-only LLM 预训练中的稀疏 MoE，重点覆盖 FFN MoE，同时讨论会反向影响结构选择的 Expert Parallel（EP）、All-to-All、token capacity 与 expert placement。多模态 MoE、MoE-LoRA、外部模型 ensemble 和纯后训练专家融合不在本文范围内。模型能力数字只用于说明架构可扩展性，不能作为跨论文的因果比较：训练数据、token 数、优化器、上下文长度、post-training 和评测污染控制往往并不一致。

\section{统一形式化：MoE 同时优化什么}

\subsection{Token-choice MoE}

给定第 $t$ 个 token 的隐藏表示 $x_t\in\mathbf{R}^{d}$，Router 计算 expert affinity：
\begin{equation}
  s_{i,t}=\phi(x_t,e_i),\qquad
  \mathcal{K}_t=\operatorname{TopK}_{i\in\{1,\ldots,N\}}(s_{i,t}),
  \label{zh:eq:router}
\end{equation}
其中 $N$ 是 routed experts 数，$e_i$ 是 expert routing embedding 或 Router 权重。输出为
\begin{equation}
  y_t=x_t+\sum_{i\in\mathcal{K}_t}g_{i,t}E_i(x_t),
  \qquad
  g_{i,t}=\frac{\exp(s_{i,t})}{\sum_{j\in\mathcal{K}_t}\exp(s_{j,t})}.
  \label{zh:eq:moe}
\end{equation}
若存在 always-on shared experts，则在式~\eqref{zh:eq:moe}中增加 $\sum_j E^{\mathrm{shared}}_j(x_t)$。从公式看，MoE 只是稀疏函数组合；从系统看，$\mathcal{K}_t$ 决定 token 要跨越哪些设备、每张卡获得多少 token，以及每个局部 GEMM 的形状。

\subsection{容量、质量与系统效率的三目标}

MoE 并不是单目标优化。粗略地，模型希望最大化总容量 $P_{\mathrm{total}}$，同时控制单 token 激活量 $P_{\mathrm{active}}$：
\begin{equation}
  P_{\mathrm{active}} \approx P_{\mathrm{dense}}
  +\frac{k}{N}P_{\mathrm{routed}}+P_{\mathrm{shared}},
  \label{zh:eq:active}
\end{equation}
但式~\eqref{zh:eq:active}只是参数口径近似：attention、embedding、shared experts、不同 expert 宽度及框架统计方式都会造成偏差。另一方面，令一个统计窗口内 expert $i$ 接收的 token 数为 $n_i$，则负载变异系数为
\begin{equation}
  \operatorname{CV}_{\mathrm{expert}}
  =\frac{\sqrt{\frac{1}{N}\sum_i(n_i-\bar n)^2}}{\bar n}.
  \label{zh:eq:cv}
\end{equation}
低 CV 有利于 EP 吞吐，却不必然有利于知识 specialization。架构设计的根本张力是：\term{语义上允许不均匀，物理执行上又不能出现严重 straggler}。

\begin{table}[htbp]
\centering
\caption{决定 MoE 架构差异的六个问题}
\label{zh:tab:sixquestions}
\small
\begin{tabularx}{\textwidth}{p{0.25\textwidth}Y}
\toprule
设计问题 & 典型选择 \\
\midrule
谁选择谁 & token-choice、expert-choice、balanced assignment、soft merging \\
激活多少专家 & fixed Top-1/2/8、threshold、adaptive-$k$、zero-compute slot \\
Expert 多大 & coarse homogeneous、fine-grained、tiny、heterogeneous size \\
是否有公共路径 & pure routed、shared expert、dense residual、cross-layer shortcut \\
在哪个范围均衡 & sequence、micro-batch、global batch、device、node、communication \\
谁处理运行时动态性 & auxiliary loss、router bias、capacity/drop、placement、结构保证 \\
\bottomrule
\end{tabularx}
\end{table}

\subsection{四个控制面：对象、决策、控制与执行}

表\ref{zh:tab:sixquestions}中的设计问题并不处于同一层次。为了区分“历史上架构如何演进”和“一个具体 MoE 系统内部如何工作”，本文进一步按\term{状态变量、决策粒度和更新时间尺度}把系统横截面分为四个控制面：Topology 是较慢变化的模型结构；Routing 是逐 token 的离散决策；Balance 在一组 tokens 上统计并施加反馈；Expert Parallel（EP）把逻辑决策落实为设备上的通信与 kernel。第三章将沿历史时间轴讨论瓶颈迁移，第四至七章则分别展开这四个控制面；\textbf{后者不是四个新的发展阶段。}

形式上，可以把四层关系写成
\begin{align}
  \mathcal{E} &= T(\theta_{\mathrm{topo}}),
  &\text{Topology：构造 expert 集合、分组和共享关系}; \notag\\
  \mathcal{K}_t &= R(x_t,\mathcal{E};\theta_{\mathrm{route}},b),
  &\text{Routing：为 token $t$ 选择 expert 子集}; \notag\\
  n_i &= \sum_t \mathbf{1}[i\in\mathcal{K}_t],\quad
  (b,\alpha,c)\leftarrow C(\{n_i\},\pi),
  &\text{Balance：依据聚合负载更新 bias、loss 或 capacity}; \notag\\
  y_t &= \operatorname{EP}\!\left(x_t,\mathcal{K}_t,\pi,\sigma\right),
  &\text{EP：按 placement $\pi$ 和 schedule $\sigma$ 执行 dispatch/combine}. \label{zh:eq:fourplanes}
\end{align}
其中 $\mathcal{E}$ 是逻辑 expert 集合，$\mathcal{K}_t$ 是 token 的路由结果，$n_i$ 是统计窗口内 expert $i$ 的负载，$b$、$\alpha$、$c$ 分别代表 Router bias、辅助损失强度和 capacity 控制量，$\pi$ 是 expert 到设备的映射，$\sigma$ 是通信与计算调度。式\eqref{zh:eq:fourplanes}说明：四层并非独立模块，而是一个闭环。

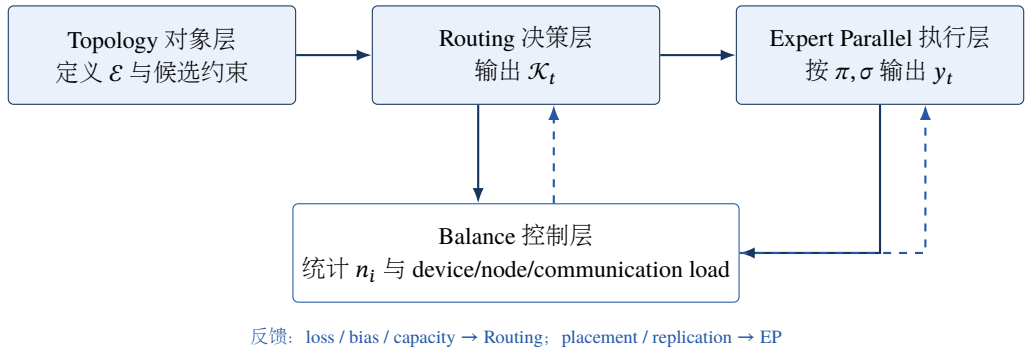
\begin{figure}[htbp]
\centering
\begin{tikzpicture}[
  node distance=10mm and 10mm,
  plane/.style={draw=DeepBlue,rounded corners=2pt,fill=SoftBlue,minimum width=38mm,minimum height=13mm,align=center,font=\small},
  control/.style={draw=Accent,rounded corners=2pt,fill=white,minimum width=58mm,minimum height=13mm,align=center,font=\small},
  arr/.style={-{Latex[length=2mm]},thick,draw=DeepBlue},
  fb/.style={-{Latex[length=2mm]},thick,dashed,draw=Accent}
]
\node[plane] (topo) {Topology 对象层\\定义 $\mathcal{E}$ 与候选约束};
\node[plane,right=of topo] (route) {Routing 决策层\\输出 $\mathcal{K}_t$};
\node[plane,right=of route] (ep) {Expert Parallel 执行层\\按 $\pi,\sigma$ 输出 $y_t$};
\node[control,below=13mm of route] (bal) {Balance 控制层\\统计 $n_i$ 与 device/node/communication load};
\draw[arr] (topo)--(route);
\draw[arr] (route)--(ep);
\draw[arr] ([xshift=-5mm]route.south)--([xshift=-5mm]bal.north);
\draw[fb] ([xshift=5mm]bal.north)--([xshift=5mm]route.south);
\draw[arr] (ep.south) |- (bal.east);
\draw[fb] (bal.east) -| ([xshift=6mm]ep.south);
\node[font=\scriptsize\color{Accent},below=2mm of bal] {反馈：loss / bias / capacity $\rightarrow$ Routing；placement / replication $\rightarrow$ EP};
\end{tikzpicture}
\caption{Topology、Routing、Balance 与 Expert Parallel 的闭环关系。前三条实线形成前向执行链，虚线表示统计与系统代价对路由和放置的反馈。}
\label{zh:fig:fourplanes}
\end{figure}

\begin{table}[htbp]
\centering
\caption{四个控制面的划分标准}
\label{zh:tab:fourplanes}
\small
\begin{tabularx}{\textwidth}{>{\raggedright\arraybackslash}p{0.17\textwidth}>{\raggedright\arraybackslash}p{0.22\textwidth}>{\raggedright\arraybackslash}p{0.25\textwidth}Y}
\toprule
控制面 & 核心问题 & 主要状态与时间尺度 & 典型失败模式 \\
\midrule
Topology 对象层 & 有哪些 experts，它们多大、如何分组或共享 & 权重结构；架构设计/训练期慢变量 & 知识冗余、粒度过粗、共享路径过重 \\
Routing 决策层 & 当前 token 应选择哪些 experts、激活多少 & affinity、Top-$k$、候选约束；逐 token 快变量 & misrouting、离散优化困难、语义受拓扑约束 \\
Balance 控制层 & 一组 tokens 的使用分布是否满足训练和系统目标 & expert/device/node load；sequence 到 runtime 多尺度反馈 & collapse、过度均匀、straggler、通信热点 \\
Expert Parallel 执行层 & 选中的 expert 在何处、以何种通信和 kernel 执行 & placement、dispatch/combine、schedule；step/runtime 变量 & All-to-All 暴露、小 GEMM、P99 延迟、显存热点 \\
\bottomrule
\end{tabularx}
\end{table}

四层之间存在两类不能忽略的反向依赖。第一，物理拓扑会约束语义路由：device-limited routing 为降低 fan-out，主动缩小 $\mathcal{E}$ 中对某个 token 可见的候选集合。第二，Balance 不只调 Router：runtime replication 或 expert placement 可以在不改变 $\mathcal{K}_t$ 的情况下改善物理负载。因此“负载均衡”不能只理解为一个 auxiliary loss，“Expert Parallel”也不是模型结构确定后才介入的被动实现。

\section{八个演进节点：划分标准、主干与分支}

这里的“八个”不是从某篇 Survey 直接摘录的固定 taxonomy，也不是按年份机械切段，而是本文依据\term{主导瓶颈迁移}做出的分析性归纳。一个变化只有同时满足以下三点，才被单列为演进节点：第一，它引入了可独立调节的新设计变量，例如 Top-$k$、expert 粒度、shared path 或动态 active budget；第二，它使系统的主要矛盾发生迁移，例如从计算随 expert 数线性增长，转为离散路由与负载不均，再转为 All-to-All、小 GEMM 或权重 I/O；第三，它被后续架构继承，因而不是某个模型的一次性实现细节。

按这一标准，节点 1--6 构成较清晰的历史主干：统计分工 $\rightarrow$ 稀疏激活 $\rightarrow$ Transformer 规模化 $\rightarrow$ decoder-only 产品化 $\rightarrow$ 细粒度知识组织 $\rightarrow$ ultra-sparse 容量扩展。节点 7 和节点 8 则不是简单接在节点 6 之后的新“代际”：节点 7 放松“每个 token 固定计算量”的假设；节点 8 放松“语义路由、层内 expert 拓扑和物理通信必须绑定”的假设。二者都可与节点 5 或 6 的 expert 结构组合。图\ref{zh:fig:evolution-dag}给出这种继承关系。

\begin{figure}[htbp]
\centering
\begin{tikzpicture}[
  node distance=5mm and 5mm,
  stage/.style={draw=DeepBlue,rounded corners=2pt,fill=SoftBlue,minimum width=37mm,minimum height=11mm,align=center,font=\small},
  branch/.style={draw=Accent,rounded corners=2pt,fill=white,minimum width=52mm,minimum height=11mm,align=center,font=\small},
  arr/.style={-{Latex[length=2mm]},thick,draw=DeepBlue},
  darr/.style={-{Latex[length=2mm]},thick,dashed,draw=Accent}
]
\node[stage] (s1) {1 统计分工\\Dense/Soft MoE};
\node[stage,right=of s1] (s2) {2 稀疏激活\\Top-$k$ conditional compute};
\node[stage,right=of s2] (s3) {3 集群规模化\\Transformer + EP};
\node[stage,below=8mm of s3] (s4) {4 产品化\\开放 decoder MoE};
\node[stage,left=of s4] (s5) {5 知识细分\\Fine-grained + shared};
\node[stage,left=of s5] (s6) {6 容量继续扩展\\Ultra-sparse scaling};
\node[branch,below=9mm of s6,xshift=14mm] (s7) {7 计算预算动态化\\adaptive-$k$ / zero-compute experts};
\node[branch,below=9mm of s4,xshift=-14mm] (s8) {8 语义与物理解耦\\ScMoE / 异构 / 跨层 / 结构化通信};
\draw[arr] (s1)--(s2);
\draw[arr] (s2)--(s3);
\draw[arr] (s3)--(s4);
\draw[arr] (s4)--(s5);
\draw[arr] (s5)--(s6);
\draw[darr] (s5.south west)--(s7.north);
\draw[darr] (s6.south)--(s7.north west);
\draw[darr] (s5.south east)--(s8.north west);
\draw[darr] (s4.south)--(s8.north east);
\end{tikzpicture}
\caption{八个演进节点的关系。实线表示历史主干上的主要继承，虚线表示可叠加的正交分支；节点编号不代表严格代际替换。}
\label{zh:fig:evolution-dag}
\end{figure}

\begin{table}[htbp]
\centering
\caption{八个节点的划分依据与瓶颈迁移}
\label{zh:tab:evolution-nodes}
\small
\begin{tabularx}{\textwidth}{p{0.055\textwidth}p{0.18\textwidth}p{0.25\textwidth}Y}
\toprule
节点 & 新增设计变量 & 解决的主导问题 & 转移出的新瓶颈 \\
\midrule
1 & gating 与 expert specialization & 单模型难以对输入空间分区建模 & 所有 experts 都计算，容量与 FLOPs 绑定 \\
2 & 稀疏 Top-$k$ 与 capacity & total parameters 与 active FLOPs 解耦 & 离散路由、collapse、token drop \\
3 & EP、All-to-All、稳定化 loss & Transformer MoE 的集群级训练 & 通信、straggler、数值与负载稳定性 \\
4 & decoder-only 训练与部署闭环 & 证明经典 MoE 可进入开放模型生态 & 大 expert 知识冗余、组合粒度粗 \\
5 & fine-grained、shared、device limit & 公共知识重复与专家组合不足 & fan-out、小 GEMM、共享路径固定成本 \\
6 & 更大 $N$、tiny experts、检索式路由 & 固定 active budget 下继续扩 total capacity & 权重 I/O、检索、低 token/expert \\
7 & adaptive compute / zero slots & token 难度不同却使用固定 Top-$k$ & active budget 控制、尾延迟、训练稳定性 \\
8 & shortcut、异构/跨层拓扑、结构化通信 & 语义路由被设备拓扑和通信绑定 & 调度复杂度、缓存一致性、大规模验证 \\
\bottomrule
\end{tabularx}
\end{table}

\subsection{Dense/Soft MoE：统计分工而非计算稀疏}

Jacobs 等人在 1991 年提出 gating network 与 local experts 的基本形式\cite{zh:jacobs1991adaptive}。这一节点建立的是\term{统计分工}：gate 根据输入产生混合权重，不同子网络拟合输入空间的不同区域，训练目标可以推动 specialization。由于所有 experts 通常都参与加权，路由是连续可导的，既不需要 token capacity，也不存在 All-to-All 意义上的稀疏 dispatch。

\textbf{为什么单列。}它定义了后来 MoE 一直保留的三个角色：Router/gate、experts 和 weighted combine。\textbf{为何还不是现代稀疏 MoE。}增加 expert 数会近似线性增加计算，因此“模型容量”和“每 token FLOPs”尚未解耦。节点 2 继承其统计分工目标，但将 dense mixture 改为只执行少数 experts；这一步才产生 conditional scaling，也同时产生离散选择带来的训练与系统问题。

\subsection{Sparse conditional computation：Top-k 建立容量杠杆}

Shazeer 等人将 noisy Top-$k$ routing 用于超大稀疏网络\cite{zh:shazeer2017outrageously}，使每个输入只执行少数 experts。若总 expert 数为 $N$、每个输入只激活 $k\ll N$ 个 experts，那么 total capacity 可以随 $N$ 增长，而主要 expert FLOPs 近似只随 $k$ 增长。这里第一次形成现代 MoE 的容量杠杆：\term{参数容量与单 token 计算量解耦}。

这一解耦并非免费。Top-$k$ 使未选 expert 没有来自该 token 的梯度；热门 experts 会溢出 capacity，冷门 experts 可能长期得不到训练；跨设备执行还必须加入 dispatch/combine。因而 noisy routing、importance/load auxiliary loss、capacity factor 和 token drop 不是外围技巧，而是稀疏执行本身诱发的配套机制。节点 3 并不改变这一基本算法，而是回答：当 MoE 被反复嵌入 Transformer 层、扩展到数千设备时，如何让它稳定且可执行。

\subsection{Transformer MoE 与 Expert Parallel}

GShard 将 Top-2 MoE FFN 系统化嵌入 Transformer，并依靠自动 sharding 在\textbf{2048 个 TPU}上训练\textbf{超过 600B}的多语言模型\cite{zh:lepikhin2020gshard}。经典执行链由此固定为
\begin{center}
\textsf{Attention $\rightarrow$ Router $\rightarrow$ Dispatch All-to-All $\rightarrow$ Expert FFN $\rightarrow$ Combine All-to-All}.
\end{center}
这一节点的新增变量不是“更多 experts”，而是\term{expert parallelism 与集群执行语义}：token 如何跨设备重排、每个 expert 如何批处理、溢出如何处理、通信与计算如何同步。Switch Transformer 进一步采用 Top-1，以更低通信和更简单的执行换取专家组合能力与路由容错下降\cite{zh:fedus2021switch}。ST-MoE 则把稳定性提升为一等设计目标，引入 Router z-loss 约束 logits 数值尺度\cite{zh:zoph2022stmoe}。需要区分：z-loss 控制数值稳定，balance loss 控制使用分布，两者不是同一机制。

同期，BASE Layers 把训练路由写成严格均衡的线性分配问题\cite{zh:lewis2021base}，Expert Choice 则让 experts 选择固定容量 tokens\cite{zh:zhou2022expertchoice}。它们能够直接保证均衡，但 batch-level 联合分配和 expert-side capacity 对在线自回归 decoding 不如 token-choice 自然，因此没有取代 decoder-only LLM 的 Top-$k$ 主干。节点 3 的遗产是“Router + EP + capacity/balance + All-to-All”这一完整系统契约；节点 4 和 5 都沿用它，只是把重点从“能否大规模训练”转向模型质量、部署可用性和 expert 内部组织。

\subsection{开放权重 coarse-grained MoE}

Mixtral 采用\textbf{8$\times$7B、每 token Top-2}\cite{zh:jiang2024mixtral}。它没有发明新的 Router family，因此若只按算法新颖性看，似乎不应成为独立节点。但它补齐了另一条关键证据链：\term{经典 coarse-grained MoE 可以在开放权重 decoder-only LLM 中贯通预训练、指令微调、推理部署与社区复现}。

\textbf{为什么单列。}节点 3 证明的是超大集群上的训练可扩展性；节点 4 证明的是通用 decoder 模型的可用性和部署闭环。它继承 Top-$k$、同构 experts 和 EP，不改变基本执行链。也正因为系统已能稳定运行，主要矛盾才显露为知识组织问题：少量大 experts 容易重复学习公共能力，Router 只能在很粗的知识块之间组合。节点 5 正面处理这一瓶颈。

\subsection{Shared + fine-grained experts}

DeepSeekMoE 对粗粒度专家做了两项关键改造\cite{zh:dai2024deepseekmoe}：其一，把一个大 FFN 切成多个较小 experts，使 Router 在相同 active budget 下组合更多知识单元；其二，加入 shared expert isolation，把所有 token 都需要的公共知识交给 always-on 路径，减少 routed experts 的知识冗余。前者提高\term{组合分辨率}，后者把“公共能力”和“条件能力”显式拆开。DeepSeek-V2 将这一结构扩大到\textbf{236B/21B active}，并加入 device-limited routing：每个 token 的候选 experts 被限制在少量设备内，从而控制跨设备 fan-out\cite{zh:deepseekv2}。

共享或稠密路径还能提供隐藏通信的计算窗口。DeepSeek-V2 将 shared expert computation 与 EP communication overlap；Snowflake Arctic 的 dense residual path 也体现了同类算法--系统协同\cite{zh:snowflakearctic}。但\term{shared expert 不是 fine-grained MoE 的必要条件}。Qwen3-235B-A22B 采用\textbf{128 个细粒度 experts、Top-8}，却移除 shared expert，并依靠 global-batch balance 允许常用能力在 routed experts 中自然形成\cite{zh:qwen3}。

因此，节点 5 实际包含两个可拆分的轴：expert granularity 与 shared path。它们经常组合，但不存在逻辑绑定。节点 5 继承节点 3 的 EP 系统，却把 expert 数、Top-$k$ 和设备 fan-out 同时推高，使小 GEMM、路由碎片化和通信拓扑成为新瓶颈。节点 6 沿着“更多、更小 experts”继续扩展；节点 8 则试图改变 fan-out 与通信的物理实现。

\subsection{Ultra-sparse scaling}

这一阶段的问题变为：\emph{固定 active parameters 时，继续增加 total experts 是否仍能降低 loss？}PEER 用 product-key retrieval 从\textbf{超过百万个 tiny experts}中选择少数 experts\cite{zh:he2024peer}；Kimi K2 将工程可落地的 ultra-sparse 结构扩展到\textbf{1.04T/32.6B active、384 routed experts、Top-8}，并报告固定 active budget 下更高 sparsity 的 scaling 收益\cite{zh:kimiK2}；gpt-oss-120b 则采用\textbf{128 experts、Top-4}和较小 active footprint，体现面向本地部署的高稀疏路线\cite{zh:openaiGptOss}。

\textbf{为什么从 fine-grained 单列。}fine-grained 的首要目标是减少知识冗余、提高组合分辨率；ultra-sparse 的首要问题则是 scaling：固定 active parameters 和训练 FLOPs 时，扩大候选 expert 集是否持续改善 loss。二者都增加 expert count，但优化目标不同。

\term{稀疏度不能无限增加。}随着 $N$ 增大，Router scoring、expert weight I/O、All-to-All metadata 和小 GEMM 碎片化都会上升；当每个 expert 每步获得的 tokens 太少时，\textbf{理论 FLOPs 优势不能转化为 wall-clock 收益}。这意味着节点 6 没有终结节点 5，反而更依赖其 fine-grained 组织和路由控制。节点 7 与 8 则从两个正交方向绕开固定 Top-$k$ 主干的限制。

\subsection{动态计算：从固定 Top-k 到 token-dependent budget}

固定 Top-$k$ 隐含所有 token 使用相同数量的 expert FLOPs，即便 token 的语义难度和所需容量不同。LongCat-Flash 加入 zero-computation experts：Router 仍选择固定 slots，但部分 slots 是 identity，因此真实执行的 FFN 数量随 token 变化\cite{zh:longcatFlash}。adaptive-$k$、threshold routing 与 zero-compute slots 的共同目标，是\term{把“选哪个 expert”和“到底计算多少”从一个固定超参数中拆开}。

\textbf{与节点 6 的关系。}ultra-sparse 扩大的是候选容量 $N$，但通常仍保持固定 $k$；动态计算改变的是每个 token 的实际 $k_t$ 或有效 FFN 数。两者可以组合：候选池很大，同时简单 token 少算、困难 token 多算。控制目标也由单一 expert load 扩展为“平均 active budget + budget variance + real-expert load + P99 latency”。训练时的平均 FLOPs 约束并不能自动保证在线尾延迟，因为困难 tokens 可能在同一 micro-batch 或请求窗口内聚集。

动态计算的遗留问题是闭环控制：若 Router 同时决定语义 expert 和计算预算，balance bias 既影响 specialization，又影响吞吐；若只约束长期均值，又可能出现短时 burst。因而 adaptive bias/PID 式反馈、budget loss 与 runtime admission control 是配套问题，而非单纯修改 Top-$k$ 即可解决。

\subsection{语义路由与物理执行解耦}

节点 1--7 主要改变 Router 可选的专家或 active budget；节点 8 改变的是更底层的假设：\term{语义上选中了某个 expert，并不意味着必须沿固定的层内 All-to-All 路径立即执行它}。这一节点不是单一模型家族，而是四条具有共同目标的前沿分支。

\textbf{执行依赖重排。}ScMoE 不改变 token 选哪些 experts，而是改变依赖图。其跨层 shortcut 让 routed experts 消费前层中间表示，同时用当前 dense/shared path 覆盖 dispatch/combine 的通信窗口\cite{zh:cai2024scmoe}。它解决的是 exposed communication，不直接解决 routing skew。图\ref{zh:fig:scmoe}给出抽象比较。

\begin{figure}[htbp]
\centering
\begin{tikzpicture}[
  node distance=5mm and 5mm,
  box/.style={draw=DeepBlue,rounded corners=2pt,fill=SoftBlue,minimum height=7mm,align=center,font=\small},
  comm/.style={draw=Accent,rounded corners=2pt,fill=white,minimum height=7mm,align=center,font=\small},
  arr/.style={-{Latex[length=2mm]},thick,draw=DeepBlue}
]
\node[box] (attn) {Attention / shortcut};
\node[comm,right=of attn] (dispatch) {Dispatch};
\node[box,right=of dispatch] (expert) {Routed experts};
\node[comm,right=of expert] (combine) {Combine};
\node[box,right=of combine] (merge) {Merge};
\node[box,below=8mm of expert,minimum width=35mm] (dense) {Dense/shared branch};
\draw[arr] (attn)--(dispatch);
\draw[arr] (dispatch)--(expert);
\draw[arr] (expert)--(combine);
\draw[arr] (combine)--(merge);
\draw[arr] (attn.south) |- (dense.west);
\draw[arr] (dense.east) -| (merge.south);
\node[font=\small\color{Accent},above=2mm of expert] {通信路径与 dense 计算并行};
\end{tikzpicture}
\caption{ScMoE 的核心是重排执行依赖以扩大通信隐藏窗口；它不是新的路由算法。}
\label{zh:fig:scmoe}
\end{figure}

LongCat-Flash 将 ScMoE 与 token-dimension chunking、Single Batch Overlap 结合；LongCat-2.0 的官方报告进一步在专用 accelerator 上采用 dense/MoE per-core full parallel execution\cite{zh:longcat2}。这些进展降低 exposed communication，\textbf{但不会自动解决严重 routing skew、expert 内部计算差异和显存压力}。

\textbf{异构 expert。}MoHGE 采用两级路由：先选 expert group，再在组内选择不同大小 experts，并结合 group-wise 与 intra-group balance 约束部署\cite{zh:mohge2026}。它把“expert 等宽”从默认约束变成设计变量，使不同 token 可获得不同形态的计算能力；代价是异构 GEMM、容量规划和负载指标都更复杂。

\textbf{跨层共享。}GMoE 把一部分 experts 设为跨层共享的 Global Experts，同时保留 layer-local experts，以减少跨层能力冗余\cite{zh:gmoe2026}。这与 shared expert 的区别在于共享作用域：前者跨 layer，后者通常是在单层内对所有 tokens always-on。跨层共享会引入缓存、调度和层间干扰问题。

\textbf{结构化通信。}Multi-Head LatentMoE 与 Head Parallel 尝试改变通信拓扑，使通信量相对 activated expert 数 $k$ 为 $O(1)$，并获得更确定的流量\cite{zh:cui2026latentmoe}。与 ScMoE“隐藏动态通信”不同，这条路线试图“减少或结构化通信本身”。由于尚缺少 trillion-parameter、万卡级公开验证，\term{它应被视为研究前沿，而非已经替代 Top-$k$ EP 的主流}。

四条分支分别优化时间重叠、计算形态、参数复用和数据移动，可组合而非替代。

\section{Expert topology：对象层}

在第二章的四层框架中，Topology 是对象层：它先定义 expert 集合 $\mathcal{E}$、知识单元的粒度、共享范围和计算异质性，Routing 才能在其上做逐 token 选择。因而，Topology 的核心不只是 expert count，而是\term{Router 可以组合什么样的计算单元}。

把 MoE 演进只理解为 expert count 增长是不充分的。表\ref{zh:tab:topology}显示，真正变化的是知识单元的粒度、共享范围和计算异质性。

\begin{table}[htbp]
\centering
\caption{Expert topology 的主要演进路线}
\label{zh:tab:topology}
\small
\begin{tabularx}{\textwidth}{p{0.21\textwidth}p{0.23\textwidth}Y Y}
\toprule
拓扑 & 代表工作 & 主要收益 & 新瓶颈 \\
\midrule
少量同构大专家 & GShard、Switch、Mixtral & 实现 conditional scaling & 知识重复、组合性有限 \\
Shared/dense + routed & DeepSeekMoE、Arctic & 公共知识隔离，可制造 overlap window & 固定 FLOPs 增加，公共路径可能过重 \\
Fine-grained experts & DeepSeek-V2/V3、Qwen3 & 更细知识组合，更高 total/active ratio & 路由、通信与 GEMM 更碎 \\
Ultra-sparse experts & PEER、Kimi K2、gpt-oss & 固定 FLOPs 下继续扩容量 & 检索、权重 I/O、低 token/expert \\
Heterogeneous/dynamic & LongCat、MoHGE & token 级动态计算预算 & budget control 与物理均衡耦合 \\
Cross-layer global/local & GMoE & 减少跨层 expert 冗余 & 调度、缓存与层间干扰待验证 \\
Latent/head structured & Multi-Head LatentMoE & 通信确定化、与 $k$ 解耦 & 大规模质量证据不足 \\
\bottomrule
\end{tabularx}
\end{table}

另一个经常被忽略的维度是初始化方式。Sparse Upcycling 将已有 dense FFN 复制成多个 experts，再新增 Router 继续训练\cite{zh:komatsuzaki2022upcycling}。它复用 dense checkpoint，却带来 expert symmetry，需要路由噪声、数据差异和后续训练使 experts 分化。更进一步的 Expert Upcycling 从已有 $E$-expert MoE 扩展到 $mE$ experts，同时保持 Top-$k$ 与 active cost 不变\cite{zh:expertUpcycling2026}。upcycling 不是新的 Router family，但会改变 specialization 的形成路径和训练成本。

\section{Routing：决策层}

在图\ref{zh:fig:fourplanes}的闭环中，Routing 接收 Topology 定义的 expert 集合与系统施加的候选约束，对每个 token 产生 $\mathcal{K}_t$。它优化的是\term{局部语义匹配}，并不直接保证一个 sequence、batch 或 device 上的总体负载合理；后者属于第六章的控制层。

\begin{table}[htbp]
\centering
\caption{主要 Router family 及其适用边界}
\label{zh:tab:router}
\small
\begin{tabularx}{\textwidth}{p{0.24\textwidth}p{0.19\textwidth}Y Y}
\toprule
Router family & 决策单位 & Balance 特性 & decoder-only LLM 适用性 \\
\midrule
Noisy Top-$k$ token-choice & token 选 experts & 需 loss、bias 或 runtime 均衡 & 当前主干，在线自然 \\
Top-1 Switch & token 选 1 expert & 通信简单，capacity 更敏感 & 可扩展，但组合能力受限 \\
Balanced assignment & batch 联合分配 & 训练时严格等量 & 全局求解复杂，在线不自然 \\
Expert-choice & expert 选 tokens & expert 容量天然固定 & encoder 更自然，decoder capacity 难处理 \\
Soft/parameter merging & 连续混合 & 可微、无离散 Top-$k$ & 大规模 EP 效率尚未证明 \\
Bias-controlled Top-$k$ & affinity + 非梯度 bias & 不把均衡梯度写入 LM 参数 & DeepSeek-V3 等模型采用 \\
Zero experts/adaptive-$k$ & token 计算量可变 & 需约束平均预算与尾部 & 动态计算的重要方向 \\
\bottomrule
\end{tabularx}
\end{table}

Token-choice 成为主流并不表示它在优化上最优，而是它最符合自回归在线约束：当前 token 的路由只依赖当前状态，不需要等待未来 tokens 或求解整个 batch 的全局分配。Soft MoE 和 Lory 等连续路由方案避免离散 Top-$k$，但尚未在超大规模 decoder 与低延迟 EP 上证明系统优势\cite{zh:puigcerver2023softmoe,zh:zhong2024lory}。

路由自由度还受到网络拓扑约束。device-limited 或 node-limited routing 将候选 expert 限制在少量设备/节点集合，降低单 token 的通信 fan-out。代价是语义最优 expert 可能被拓扑约束排除，因此这类设计本质上是质量与通信之间的显式折中。

\section{Balance：控制层}

Balance 不重新定义 experts，也不独立生成 token 的语义路由。它观察许多次 Routing 决策累积出的 $\{n_i\}$ 及设备、节点和通信负载，再通过 loss、bias、capacity 或 runtime placement 反馈给决策层和执行层。其核心问题是：\term{允许语义 specialization 不均匀的同时，如何避免训练 collapse 和物理 straggler}。

\subsection{Sequence/batch-wise expert balance}

经典辅助损失可写为
\begin{equation}
  \mathcal{L}_{\mathrm{bal}}=\alpha N\sum_{i=1}^{N} f_i P_i,
  \label{zh:eq:balance}
\end{equation}
其中 $f_i$ 是统计窗口内实际分配到 expert $i$ 的 token fraction，$P_i$ 是平均 routing probability，$\alpha$ 控制均衡强度。当所有 experts 使用近似均匀时，式~\eqref{zh:eq:balance}较小。该损失同时承担防止 expert collapse 和避免 EP straggler 两项职责，但二者并不等价：模型可能需要不均匀 specialization，硬件却希望均匀负载。$\alpha$ 过大时，硬件目标会直接干扰语言建模梯度。

\subsection{Device/node/communication-aware balance}

DeepSeek-V2 把 balance 拆为 expert-level、device-level 与 communication balance，并通过 device-limited routing 控制 fan-out\cite{zh:deepseekv2}。这一变化的关键不是多加几个 loss，而是承认\term{逻辑 expert 均衡不等于物理设备或网络均衡}：若一个 device 上有多个 experts，只要 device 间总 token 接近，未必需要所有 experts 完全等量。

\subsection{Auxiliary-Loss-Free Load Balancing}

Auxiliary-Loss-Free（ALF）策略给每个 expert 增加非梯度 routing bias $b_i$：
\begin{equation}
  \mathcal{K}_t=\operatorname{TopK}_i(s_{i,t}+b_i),\qquad
  b_i\leftarrow b_i+\eta\,\operatorname{sign}(\bar n-n_i),
  \label{zh:eq:alf}
\end{equation}
其中 $\eta$ 是 bias 更新速度。bias 影响 Top-$k$ 选择，但不作为 affinity 权重参与 LM 梯度，从而减少均衡目标对模型参数更新方向的直接干扰\cite{zh:wang2024auxfree}。DeepSeek-V3 使用 batch-wise ALF 作为主均衡机制，同时保留很弱的 sequence-wise auxiliary loss，防止单序列极端失衡\cite{zh:deepseekv3}。因此，\term{“DeepSeek-V3 完全没有任何 balance loss”并不准确}。

global-batch 统计比单 micro-batch 更稳定，也允许某些 experts 在局部 batch 上承载更多特定 domain token；代价是需要跨 DP/EP 范围聚合负载，扩大统计范围可能引入额外通信。

\subsection{Runtime expert placement}

训练分布上的均衡不能保证真实推理流量。domain shift、SFT/RL 和关闭 balance 更新都会改变 expert popularity。DeepSeek-V3 使用冗余 expert deployment，在运行时复制热点 experts 并调整 placement\cite{zh:deepseekv3}。这一趋势把目标重新分工：Router 可以保留语义上的非均匀 specialization，runtime 通过复制、迁移或重新放置避免设备 straggler。

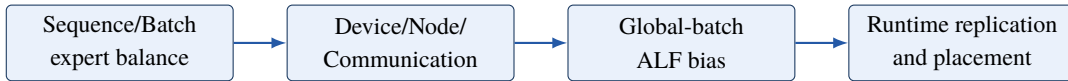
\begin{figure}[htbp]
\centering
\begin{tikzpicture}[
  stage/.style={draw=DeepBlue,fill=SoftBlue,rounded corners=2pt,minimum width=30mm,minimum height=10mm,align=center,font=\small},
  arr/.style={-{Latex[length=2mm]},thick,draw=Accent}
]
\node[stage] (s1) {Sequence/Batch\\expert balance};
\node[stage,right=7mm of s1] (s2) {Device/Node/\\Communication};
\node[stage,right=7mm of s2] (s3) {Global-batch\\ALF bias};
\node[stage,right=7mm of s3] (s4) {Runtime replication\\and placement};
\draw[arr] (s1)--(s2);
\draw[arr] (s2)--(s3);
\draw[arr] (s3)--(s4);
\end{tikzpicture}
\caption{Balance 的演进不是简单减弱约束，而是逐步把物理执行目标从 LM 梯度中剥离。}
\label{zh:fig:balance}
\end{figure}

\section{Expert Parallel：执行层}

EP 消费 Routing 产生的 token--expert assignment，并结合 placement $\pi$ 把逻辑 expert 转化为设备上的 dispatch、局部 GEMM 与 combine。它通常不改变“语义上选中了谁”，却决定这次选择的 wall-clock 成本；当系统代价过高时，又会通过 device-limited routing、placement 或架构重构反向修改前三层。

MoE 的算法收益只有在稀疏执行高效时才成立。capacity-based 系统为每个 expert 预留固定 token slots，溢出 token 被丢弃或走 residual，容易造成训练与推理不一致。MegaBlocks 使用 block-sparse kernels 实现 dropless MoE，使动态 token shape 不再要求 token dropping\cite{zh:gale2022megablocks}。Tutel 等系统则动态选择并行和 pipeline 策略，以适配不同集群规模\cite{zh:hwang2022tutel}。

随着专家更细、Top-$k$ 更大，通信逐渐成为一等架构约束，推动出三种处理方式：
\begin{enumerate}
  \item \term{限制通信拓扑}：device/node-limited routing 降低 fan-out；
  \item \term{隐藏通信}：shared computation overlap、DualPipe、ScMoE 和 token chunking 扩大 overlap window；
  \item \term{重构通信}：Head Parallel 等方案把动态流量变为更确定的通信模式。
\end{enumerate}

这里\term{必须区分“总 All-to-All 时间”和“exposed communication time”}。若通信完全被 dense path 覆盖，继续优化链路带宽对端到端 step time 的边际收益会下降；反之，即使总通信量不变，只要关键路径缩短，吞吐也可能显著改善。因此训练报告至少应同时给出\textbf{通信量、overlap 比例、exposed dispatch/combine、MFU 和 step-time P99}。

\section{现代主干与前沿分支}

截至 2026 年，大规模 decoder MoE 最常见的组合可概括为：
\begin{center}
\fcolorbox{DeepBlue}{SoftGray}{\parbox{0.88\textwidth}{\centering
Token-choice Top-$k$ + fine-grained experts + 可选 shared path + global/batch-wise balance 或 ALF + topology-limited routing + dropless kernels/communication overlap + runtime expert placement}}
\end{center}

DeepSeek-V3、Qwen3 和 Kimi K2 在 shared expert、balance 机制和 EP size 上不同，但都位于这条主干。与此同时，以下分支仍未形成统一结论：
\begin{itemize}
  \item \term{MoE Attention}：JetMoE 同时稀疏化 Attention 与 FFN，进一步降低 active FLOPs，但 KV/cache 和路由更复杂\cite{zh:shen2024jetmoe}；
  \item \term{Hybrid mixer + MoE}：Jamba 在 Transformer 与 Mamba layers 间交替，说明 MoE 是参数扩容方式，不绑定全 Attention backbone\cite{zh:lieber2024jamba}；
  \item \term{Million experts}：理论组合性高，但权重 I/O、小 GEMM 和检索开销仍限制工业部署；
  \item \term{Heterogeneous experts}：按 token 难度选择不同容量，表达自然，但物理放置和尾延迟更难；
  \item \term{Cross-layer global experts}：目标是消除层间重复能力，仍缺少超大规模训练证据；
  \item \term{Structured communication}：Head Parallel 可能降低动态 EP 痛点，但尚处早期验证阶段。
\end{itemize}

表\ref{zh:tab:models}以架构选择而非 benchmark 排名对代表性模型进行对照。

\begin{table}[htbp]
\centering
\caption{代表性 LLM MoE 的架构位置；active 参数口径因报告而异}
\label{zh:tab:models}
\scriptsize
\begin{tabularx}{\textwidth}{p{0.18\textwidth}C{15mm}C{25mm}C{22mm}Y}
\toprule
模型 & 年份 & Total/Active & Experts/Top-$k$ & 架构意义 \\
\midrule
GShard & 2020 & $>$600B/未统一披露 & Top-2 & Transformer MoE 与自动 sharding \\
Switch & 2021 & up to 1T+ & Top-1 & 用简单路由换吞吐与可扩展性 \\
Mixtral 8$\times$7B & 2024 & 46.7B/12.9B & 8/2 & 开放 coarse-grained MoE 标杆 \\
DeepSeek-V2 & 2024 & 236B/21B & 160/6 & fine-grained、shared、device-limited \\
DeepSeek-V3 & 2024 & 671B/37B & 256/8 & ALF、weak seq loss、node-limited \\
Qwen3-235B-A22B & 2025 & 235B/22B & 128/8 & fine-grained、无 shared、global balance \\
Kimi K2 & 2025 & 1.04T/32.6B & 384/8 & ultra-sparse scaling \\
gpt-oss-120b & 2025 & 116.8B/5.1B & 128/4 & 小 active footprint 与部署导向 \\
LongCat-Flash & 2025 & 560B/18.6--31.3B & real + zero & 动态计算预算与 ScMoE \\
LongCat-2.0 & 2026 & 1.6T/33--56B & dynamic & dense/MoE per-core 并行 \\
\bottomrule
\end{tabularx}
\end{table}

\section{面向基座预训练的实验设计}

\term{跨模型 benchmark 无法识别架构贡献。}若要判断 fine-grained、shared、ALF 或 ScMoE 是否有效，应至少同时固定\textbf{训练 token、数据配比、active parameters 和优化器}，并分别报告\textbf{training-FLOPs-matched 与 wall-clock-matched}结果。

\begin{table}[htbp]
\centering
\caption{建议的等预算 MoE 消融矩阵}
\label{zh:tab:experiments}
\small
\begin{tabularx}{\textwidth}{p{0.20\textwidth}p{0.39\textwidth}Y}
\toprule
实验维度 & 建议 setting & 核心指标 \\
\midrule
Expert granularity & 固定 total/active params 与 tokens，扫描 expert count、width、Top-$k$ & val loss、下游、token/expert、GEMM efficiency \\
Shared vs emergent & shared；无 shared+global loss；无 shared+ALF & expert similarity、domain-routing MI、active FLOPs \\
Balance scope & sequence aux、global aux、ALF、ALF+weak seq & max/mean load、CV/Gini、entropy、specialization \\
Execution topology & serial EP、shared overlap、ScMoE、runtime EPLB & exposed communication、MFU、step time、TPOT、P99 \\
Dynamic compute & fixed Top-$k$、threshold、zero experts & active-param 均值/方差、分桶 loss、吞吐与尾延迟 \\
Upcycling & scratch、dense-copy、cluster/utility-aware copy & 对称性破缺速度、loss/token、GPU hours \\
\bottomrule
\end{tabularx}
\end{table}

对代码预训练场景，建议额外按语言、repo/file/function 粒度、代码与自然语言 token 分桶统计 routing。关键观测包括：不同语言的 expert co-activation、跨文件引用 token 是否集中到少数专家、代码与 code-related 文档是否共享 experts，以及 repo-level 长序列是否造成 sequence-wise load 峰值。这样才能区分“专家形成了有益 specialization”和“数据配比造成偶然流量偏斜”。

系统侧至少应同时记录：
\begin{itemize}
  \item expert、device、node 三个层级的 max/mean、CV 与 Gini；
  \item sequence、micro-batch、global batch 三个统计窗口下的负载；
  \item dispatch/combine exposed time，而不只看总 All-to-All time；
  \item token drop、capacity overflow、redundant expert hit rate；
  \item training FLOPs、active parameters、wall-clock 和集群网络拓扑。
\end{itemize}

\section{开放问题}

\subsection{Specialization 的正确分析单位是什么}

“数学专家”或“代码专家”是直观叙事，但实际路由常表现为 token 形态、语法、位置、语言和层级组合。单 expert 未必是稳定的语义单位，跨层 expert group 或 co-activation pattern 可能更可解释。未来应把 routing mutual information、expert representation similarity 和 intervention-based causal test 结合起来，而不只展示 token word cloud。

\subsection{更高 sparsity 的系统最优点在哪里}

在固定 active FLOPs 下增加 total experts 可能降低模型 loss，但系统收益受网络带宽、expert weight I/O、batch size 与 kernel arithmetic intensity 限制。最优 sparsity 不是纯模型常数，而是模型--硬件联合 scaling law。报告该结论时必须给出 EP size、网络拓扑和每 expert token 数。

\subsection{Balance 能否主要下沉到 runtime}

ALF 减少了辅助 loss 的梯度干扰，却仍会改变 Top-$k$ 选择。runtime replication 和 placement 可以允许语义路由更自由，但会带来参数副本显存、迁移成本和缓存一致性问题。训练弱约束与运行时强均衡之间的最优边界仍未确定。

\subsection{动态计算如何满足尾延迟 SLA}

zero experts 或 adaptive-$k$ 可以降低平均 FLOPs，但困难 tokens 若在同一 micro-batch 或请求窗口集中，会制造新的 straggler。模型需要同时控制平均预算、预算方差和 P99，而不能只报告平均 active parameters。

\subsection{Post-training 如何保持 Router 稳定}

SFT/RL 数据会改变领域分布和序列形态，从而改变 expert load。训练阶段是否保留 balance 更新、是否 routing replay、是否冻结 Router，不能从预训练设置直接外推。应分别测量 pretrain、SFT 和 RL 阶段的 routing drift 与 specialization retention。

\section{总结与展望}

\term{LLM MoE 的架构史不是“expert 越来越多”的单线扩张，而是五条路线的共同演化：}expert 从少量大模块变成细粒度、超稀疏和异构计算单元；拓扑从 layer-local routed experts 发展到 shared、cross-layer global/local 和 latent-head 结构；Router 从简单 Top-$k$ 逐步加入全局统计、bias 控制和动态计算预算；balance 从 sequence-wise 辅助损失走向 device/communication-aware、ALF 与 runtime placement；执行则从串行 All-to-All 发展到 dropless kernels、topology-limited routing、ScMoE overlap 和结构化通信。

因此，ScMoE 之后并不存在唯一的“下一代结构”。更可信的方向是四者组合：更自由的语义路由、显式的平均计算预算控制、拓扑感知或确定性通信，以及运行时 expert placement。架构评估也必须从单一 validation loss 扩展为质量、specialization、active budget、exposed communication 和尾延迟的联合分析。

对下一轮大规模预训练，\term{最具信息量的实验不是直接比较两个不同训练配方的终局 benchmark}，而是构造\textbf{同数据、同 token、同 active FLOPs}的受控矩阵，分别扫描 expert granularity、shared path、balance scope 和 execution topology。\textbf{只有这样，才能区分参数容量收益、路由专业化收益与系统吞吐收益}，并建立可迁移到下一代硬件的 MoE 设计结论。

\appendix
\section{架构演进速查表}

\begin{longtable}{@{}p{0.12\textwidth}p{0.19\textwidth}p{0.28\textwidth}p{0.30\textwidth}@{}}
\caption{MoE 架构演进的瓶颈迁移}\label{zh:tab:timeline}\\
\toprule
时期 & 代表工作 & 解决的主要问题 & 转移出的新瓶颈 \\
\midrule
\endfirsthead
\toprule
时期 & 代表工作 & 解决的主要问题 & 转移出的新瓶颈 \\
\midrule
\endhead
1991--2016 & Adaptive Mixtures & 学习软分工 & 计算随 expert 数线性增长 \\
2017--2019 & Sparse MoE & 参数容量与单 token FLOPs 解耦 & collapse、capacity、All-to-All \\
2020--2022 & GShard、Switch、ST-MoE & Transformer 大规模稀疏训练 & 稳定性、token drop、路由质量 \\
2023--2024 & Mixtral & 开放 decoder MoE 可用性 & 粗粒度与知识冗余 \\
2024--2025 & DeepSeekMoE/V2、Qwen3 & 细粒度组合与公共知识处理 & 通信 fan-out、小 GEMM、balance scope \\
2024--2025 & PEER、Kimi K2、gpt-oss & ultra-sparse 容量扩展 & 检索、I/O、低 token/expert \\
2024--2026 & ALF、LongCat、ScMoE & 减少梯度干扰与隐藏通信 & 运行时动态负载与尾延迟 \\
2025--2026 & MoHGE、GMoE、Head Parallel & 异构计算、跨层共享、结构化通信 & 超大规模质量和系统验证 \\
\bottomrule
\end{longtable}

\end{document}